\documentclass[10pt,twocolumn,letterpaper]{article}
\usepackage[T1]{fontenc}
\usepackage[utf8]{inputenc}
\usepackage{textcomp}

\usepackage{cvpr}              % To produce the CAMERA-READY version

\usepackage{multirow}
\usepackage{threeparttable}
\usepackage{booktabs}
\usepackage{makecell}
\usepackage[most]{tcolorbox}
\usepackage{listings}
\usepackage[ruled,vlined]{algorithm2e}
\usepackage{fvextra}
\definecolor{cvprblue}{rgb}{0.21,0.49,0.74}
\usepackage[pagebackref,breaklinks,colorlinks,allcolors=cvprblue]{hyperref}

\def\paperID{38848} % *** Enter the Paper ID here
\def\confName{CVPR}
\def\confYear{2026}

\title{LensWalk: Agentic Video Understanding by Planning How You See in Videos}

\author{
Keliang Li\textsuperscript{1,4}\quad
Yansong Li\textsuperscript{3}\quad
Hongze Shen\textsuperscript{1,2,4}\quad
Mengdi Liu\textsuperscript{1,4}\quad
Hong Chang\textsuperscript{1,4}\thanks{Corresponding author.}\quad
Shiguang Shan\textsuperscript{1,4}\\[0.8em]
\textsuperscript{1}Institute of Computing Technology, Chinese Academy of Sciences, China\\
\textsuperscript{2}Peng Cheng Laboratory, China\\
\textsuperscript{3}College of Computer Science and Electronic Engineering, Hunan University, China\\
\textsuperscript{4}University of Chinese Academy of Sciences, China\\[0.5em]
{\tt\small keliang.li@vipl.ict.ac.cn \quad liyansong@hnu.edu.cn \quad chonghong@ict.ac.cn}
}
\begin{document}
\maketitle
\begin{abstract}
The dense, temporal nature of video presents a profound challenge for automated analysis. Despite the use of powerful Vision-Language Models, prevailing methods for video understanding are limited by the inherent disconnect between reasoning and perception: they rely on static, pre-processed information and cannot actively seek raw evidence from video as their understanding evolves. To address this, we introduce LensWalk, a flexible agentic framework that empowers a Large Language Model reasoner to control its own visual observation actively. LensWalk establishes a tight reason-plan-observe loop where the agent dynamically specifies, at each step, the
temporal scope and sampling density of the video it observes. Using a suite of versatile, Vision-Language Model based tools parameterized by these specifications, the agent can perform broad scans for cues, focus on specific segments for fact extraction, and stitch evidence from multiple moments for holistic verification. This design allows for progressive, on-demand evidence gathering that directly serves the agent's evolving chain of thought. Without requiring any model fine-tuning, LensWalk delivers substantial, plug-and-play performance gains on multiple model recipes, boosting their accuracy by over 5\% on challenging long-video benchmarks like LVBench and Video-MME. Our analysis reveals
that enabling an agent to control how it sees is key to unlocking more accurate, robust, and interpretable video reasoning.
\end{abstract}    
\section{Introduction}
\label{sec:intro}
% \begin{figure}[!htb]
%     \centering
%     % \vspace{-1em}
%     \includegraphics[width=1\linewidth]{fig/intro_acc_cost_comparison.pdf}
%     \vspace{-1.6em}
%     \caption{Illustration of Accuracy and Overall Token Cost of Prior Methods and LensWalk on VideoMME long split. By making the agent plan the consumption for the video context with reasoning, LensWalk achieves the best performance and saves millions of token cost compared to existing Video Agent.}
%     % \vspace{-2em}
%     \label{fig:acc_and_cost}
% \end{figure}
Video understanding is a core task in computer vision with profound challenges, given the multi-layered, high-density, and temporally unfolding nature of its information \cite{fu2025videomme,wang2025lvbench,wu2024longvideobench}. This volume overwhelms the finite cognitive resources of any analytical agent, be it biological or artificial, making uniform, exhaustive analysis intractable and inefficient \cite{alexander2015epistemic,chandrasekharan2004reactive}. Powerful intelligence, like humans, cope through purposeful information seeking \cite{alexander2015epistemic,chandrasekharan2004reactive,parr2017uncertainty}: we continually adapt—shifting from broad peripheral scans to sharp foveal focus—and intermittently reflect and verify, guided by continual reasoning that evolves both intent and observational priorities. We contend that this principle is key to reliable, accurate video analysis: observation should be actively scheduled under limited capacity. However, realizing such purpose-oriented video observation planning in practice remains an open challenge.
\begin{figure*}[!tb]
    \centering
    % \vspace{-1em}
    \includegraphics[width=0.85\linewidth]{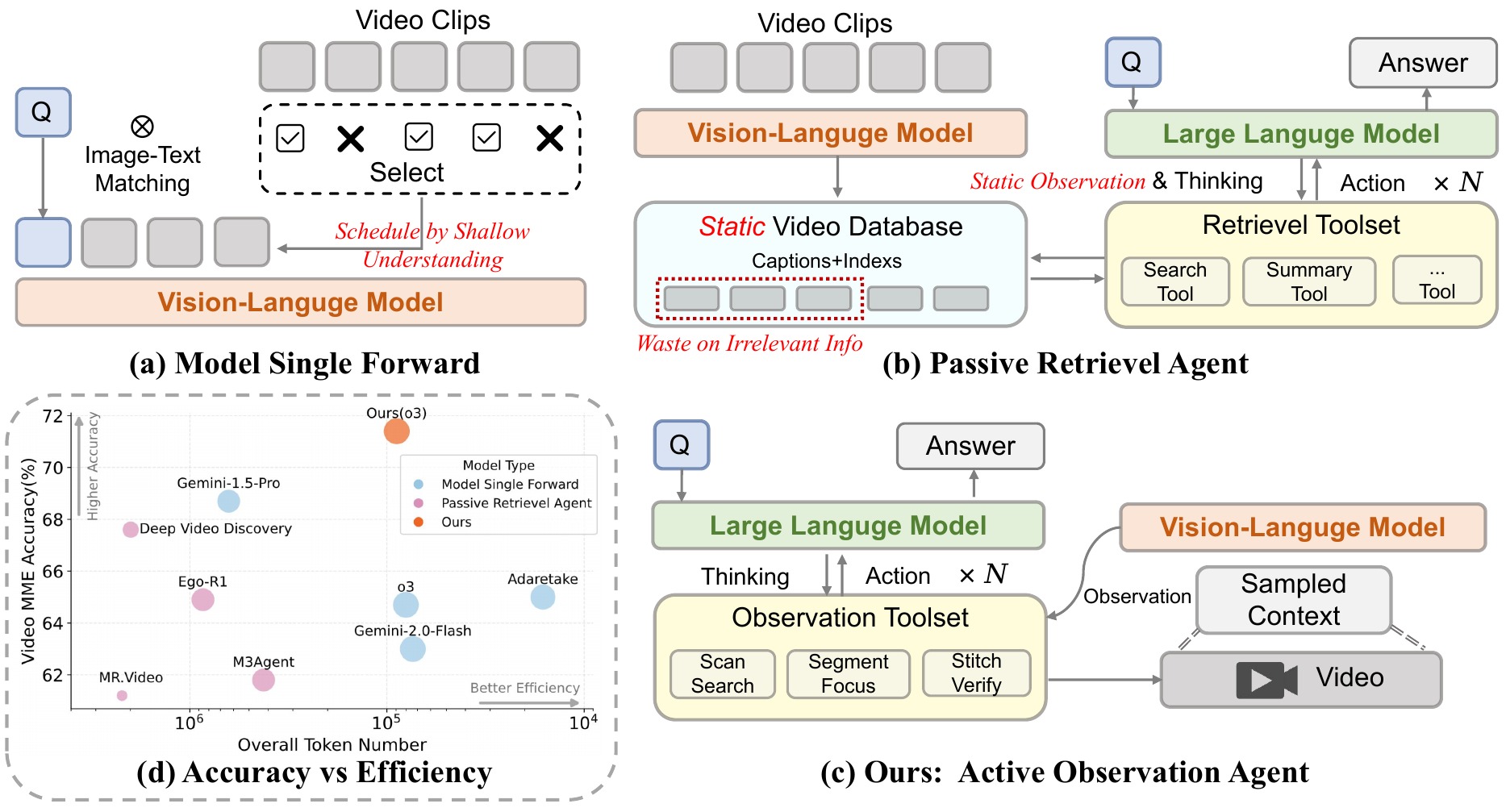}
    % \vspace{-1.6em}
    \caption{Comparison of video understanding paradigms: (a) Single Forward of model with one-off context selection. (b) Retrieval-Based Video Agent. (c) \textbf{LensWalk}, an agent framework that actively plans observations to fulfill subsequent reasoning, explores iteratively, and self-regulates its observation budget. (d) By actively planning observations only on video segments essential to its reasoning process, LensWalk simultaneously achieves high accuracy and exceptional efficiency.%, consuming significantly fewer tokens than other agentic methods.
    }
    % \vspace{-2em}
    \label{fig:compr_flow}
\end{figure*}

While the progress of Vision-Language Models (VLMs) \cite{hurst2024gpt,liu2023visual,bai2025qwen2} and test-time thinking \cite{guo2025deepseek,OpenAI2025o3o4mini} has significantly advanced the field, this principle is far from realized in current video understanding pipelines. In typical setups, long videos are sampled within context windows and converted once into a single context; this uniform strategy often drops sparse but decisive events or drowns them in redundancy \cite{wu2024longvideobench,liu2025bolt,tang2025adaptive}. More recent efforts instead try heuristic frame or token selection \cite{yu2024frame,liu2025bolt,wang2025adaretake,shen2024longvu} to preserve “key” information under a constrained context length, but these heuristics still rely on fixed, once sampling, cannot shift focus or reallocate budget as intermediate hypotheses change, and thus struggle to support the stepwise resolution of complex questions.

Beyond such single-model setups, many recent retrieval-based and agentic frameworks \cite{tian2025ego-r1,long2025seeing,pang2025mrvideo} explicitly try to expand information as reasoning unfolds. Reactive retrieval pipelines \cite{ren2025videorag,zhou2025reagent,meng2025cyberv} let the model trigger tools to fetch pre-processed artifacts such as ASR transcripts, OCR, or clip-level captions when new questions or intermediate hypotheses arise; however, perception is still bound to a largely fixed, hand-crafted workflow and a predetermined menu of views on the video, with limited ability to decide granularity or revisit the raw video when ambiguity appears. Video agents with tool flows \cite{tian2025ego-r1,long2025seeing,zhang2025deep}, as well as those that let an LLM controller query over pre-computed video indexes, offer more flexible querying but still operate on static representations rather than scheduling fresh observations from the source video, so observation and reasoning remain only loosely coupled and the overall budget remains underplanned.
% Current methodologies still treat observation and reasoning as loosely-coupled, sequential stages, not a deeply integrated process. Reactive retrieval frameworks (e.g., Video-RAG, ReAgent-V) attempt to patch this by fetching pre-processed information like subtitles or OCR. However, their perception is enslaved to a static workflow; the model can only consume from a fixed menu of information types, unable to supply evidence as fine-grained reasoning evolves.  M3Agent, and Deep Video Discovery provide more flexible querying but rely on pre-computed, static video representations. These agents navigate an indexed abstraction rather than the original visual domain, preventing them from revisiting raw data to resolve ambiguities or extract overlooked details. Moreover, because agents operate over already-refined artifacts, new observations are only weakly coupled to current demands, hindering experience building, and reflection

To this end, we present \textbf{LensWalk}, an agentic framework that gives the LLM reasoner on-hand control over where, and how dense to look at the video for active observation. LensWalk organizes inference as a multi-round \emph{reason--plan--observe} loop: at each round, the reasoner reflects on the current question and accumulated evidence, then plans an observation action by selecting an observation tool and its parameters, which determine the visible context---which temporal regions of the source video to inspect and at what sampling level. This loop is instantiated by a suite of complementary observation tools powered by a vision-language model that converts the prepared context into visual evidence: \emph{Scan Search} supports parallel sweeps over broad, disjoint candidate ranges for cue discovery, \emph{Segment Focus} constructs compact, detail-oriented views for a targeted interval, and \emph{Stitched Verify} assembles custom samples from multiple key segments into a single context for global checking. Each tool is encouraged to return frame-interleaved timestamps aligned with discovered key information as anchors for subsequent rounds. As observations are acquired progressively from raw video, LensWalk maintains a round-updated global cast table as lightweight hot memory, tracking entities and their relations across rounds without repeatedly re-reading long histories.
% To this end, we present \textbf{LensWalk}, an agentic framework that gives the LLM reasoner on-hand control over when, where, and how to look at the video, concretely instantiating purpose-oriented, budget-aware observation scheduling. LensWalk organizes inference as a multi-round reason--plan--observe process: at each round, the reasoner reflects on the current question and accumulated evidence, then plans an observation action by deciding which temporal regions of the source video to inspect, at what level of sampling, and which combination of Vision-Language Model (VLM)-powered tools to invoke. These tools prepare video context at complementary levels rather than along a fixed workflow: \emph{Scan Search} supports parallel sweeps over broad, disjoint candidate ranges for cue discovery, \emph{Segment Focus} constructs compact, detail-oriented views for a targeted interval, and \emph{Stitched Verify} assembles custom samples from multiple key segments into a single context for global checking. Each tool is encouraged to return precise timestamps aligned with discovered key information as anchors for subsequent rounds. As observations are acquired progressively from raw video, LensWalk also maintains a round-updated global cast table as lightweight hot memory, tracking entities and their relations across rounds without repeatedly re-reading long histories.
By tying each observation to an explicit, parameterized plan, LensWalk lets the agent schedule where to look and how densely to sample the video directly from its evolving reasoning state, reallocating its frame budget across regions and time scales as hypotheses change. %Unlike retrieval-based video agents that rely on pre-scanned caption or index databases, LensWalk issues fresh observations on the raw video at adaptive granularity, yielding a richer variety of observation paths and making better use of previous experience for subsequent planning. %allowing previous plans to serve as anchors and accumulated experience for subsequent planning.

% This framework lets the agent flexibly schedule observations based on its thought: it can shift focus across regions and time scales and concentrate its budget on segments most informative for its current line of reasoning. In contrast to retrieval-based video agents that rely on pre-scanned, text-centric artifacts or static indexes—incurring heavy upfront processing while remaining constrained in how and where they can refine evidence—LensWalk issues fresh observations directly on the source video, adapts granularity on demand, and uses its evolving reasoning state to balance broad exploration with focused re-inspection, thereby improving both the efficiency and effectiveness of visual evidence acquisition.
% This framework enables the agent to flexibly schedule observation of , such as shift focus between regions and time scales, and concentrate budget on the most informative segments in the agent's reasoning. In contrast to retrieval-based video agents that rely on pre-scanned, text-centric artifacts or static indexes, and thus pay the cost of upfront over-processing while remaining limited in how and where they can refine evidence, LensWalk directly schedules fresh observations from the source video, adjusts granularity on demand, and uses its evolving reasoning state to decide when to explore widely and when to re-examine specific moments in depth, substantially improving both the efficiency and effectiveness of visual evidence acquisition.

Our empirical results on long video understanding and video reasoning benchmarks reveal that this framework is not only conceptually appealing but also practically effective. As a plug-and-play agentic framework, LensWalk consistently improves even strong Reasoner–Observer combinations; when a single model serves as both to make self-scheduled observation, it effectively offers a “free lunch”, yielding substantial gains—for example, the powerful o3 model improves by over 11.5\% and 6.7\% accuracy on challenging long-video benchmarks such as LVBench and VideoMME (long split). At the same time, LensWalk exhibits an emergent ability to manage observational cost: it adaptively resolves easy queries with minimal turns, while allocating more budget only to genuinely uncertain cases, as in Figure \ref{fig:compr_flow}(d), leading to total token usage far below that of retrieval agents that consume millions of tokens and only slightly above a single dense pass over long frame sequences. Analysis of its reasoning traces further reveals diverse, human-like strategies—strategic reflection, progressive zoom-in, and integrative verification—replacing the fixed workflows or degenerative action repetition common in prior agents, and confirming that the performance and efficiency gains stem from its core principle of actively planning observation guided by reasoning.

\section{Related Work}
\label{sec:related_work}
\paragraph{Context Selection on Video Understanding}
\label{subsec:context_selection}
The constrained context window of Multimodal Large Language Models (MLLMs) has motivated a line of work on selecting a compact yet informative video context. Existing approaches typically either perform query-aware keyframe or clip selection with temporal and semantic diversity, or learn task-specific alignment modules to pick task-relevant snippets~\cite{tang2025adaptive, liu2025bolt, zhang2025q, yu2024frame, hu2025m, buch2025flexible, park2024too}. Another line focuses on information condensation by pruning redundancy at frame, clip, or model-layer granularity~\cite{yang2025pvc, shen2024longvu, song2024moviechat, wang2025adaretake, zhang2025flexselect}. While these strategies improve the utilization of the limited token budget, they still choose the visible context in a single shot before reasoning, whereas our agent treats context selection itself as a step-wise, tool-mediated process driven by its evolving hypotheses.
\vspace{-1pt} 
% The constrained context window of Multimodal Large Language Models (MLLMs) has necessitated sophisticated context selection strategies to maximize information density and minimize redundancy within the limited token budget. One primary approach involves selecting keyframes based on their relevance to a user's query, a process often along with heuristics for temporal and semantic diversity~\cite{tang2025adaptive, liu2025bolt, zhang2025q} or through dedicated trainable modules that learn task-specific alignment~\cite{yu2024frame, hu2025m, buch2025flexible, park2024too}. Another Line targets information condensation by pruning redundancy at various granularities: within individual frames~\cite{yang2025pvc}, across consecutive frames~\cite{shen2024longvu, song2024moviechat}, and even adaptively across different layers of the model, as seen in recent AdaRetake and FlexSelect\cite{wang2025adaretake, zhang2025flexselect}. While these strategies effectively enhance context window utilization, their nature as a one-time operation before the language model's inference creates a gap between the model's context focus and the static, pre-determined visual input, leaving the potential for information loss.
\paragraph{Test-Time Scaling on Video Understanding}
Recent advances in Test-Time Scaling (TTS) for video understanding allocate additional computation at inference time to improve performance. One line of work strengthens internal reasoning by training models to support long, self-corrective chains of thought with verifiable rewards or preference-based objectives, as in Video-R1 and VideoRFT~\citep{feng2025video,wang2025video,guo2025deepseek}, but still operates on fixed visual inputs. A complementary line wraps video models in agentic workflows, where LLMs control how to retrieve or aggregate preprocessed clips and captions: building on early pipeline-based systems that query clips, frames, and metadata~\citep{wang2024videoagent,suris2023vipergpt,zhou2025reagent,ren2025videorag}, recent works introduce reasoning-based agents that schedule clip retrieval and caption aggregation over long videos~\citep{zhang2025DVD,long2025seeing,tian2025ego-r1,pang2025mrvideo}. In contrast to both paradigms, our framework scales test-time inference by letting the agent explicitly decide \emph{how} to observe the video---choosing temporal ranges, sampling densities, and segment compositions---rather than only allocating more computation over a fixed observation stage.
\vspace{-2pt} 
% Recent advances in Test-Time Scaling (TTS) for video understanding aim to enhance performance by allocating additional computation at inference time, broadly falling into two paradigms: optimizing internal reasoning and deploying external agentic workflows. The first line of work focuses on unlocking long test-time chains of thought via reinforcement learning, where models are trained with verifiable rewards and preference-based optimization (e.g., GRPO) to encourage temporally consistent and self-corrective reasoning over fixed visual inputs, as in Video-R1 and VideoRFT~\citep{feng2025video,wang2025video,guo2025deepseek}. The second line leverages LLMs as video agents that interact with preprocessed video data through tool use or retrieval: building on early LLM-based systems that follow hand-crafted pipelines for querying clips, frames, and metadata~\citep{wang2024videoagent,suris2023vipergpt,zhou2025reagent,ren2025videorag}, recent frameworks employ test-time planning to automatically make decisions\cite{zhang2025DVD,long2025seeing}. For example, Ego-R1\cite{tian2025ego-r1} performs hierarchical search over long egocentric streams, while Mr.\ Video~\citep{pang2025mrvideo} iteratively aggregates information from multiple captions to refine its answers. However, although these approaches substantially scale test-time reasoning, the space of visual observations itself typically remains fixed or statically pre-indexed, so the agent’s thinking process is still constrained to passively reacting to a predetermined context.
\paragraph{Multimodal Tool-Use Agents}
Tool use is a core capability of agents, enabling models to proactively invoke external tools and thereby extend their effective context beyond intrinsic parametric knowledge~\cite{feng2025retool, zhang2025DVD, wu2025vtoolr1}. 
Recent work on multimodal tool-use agents largely follows this perspective, but instantiates it along several directions. 
One prominent line is agentic search~\cite{jiang2024mmsearch, chen2024mindsearch, jin2025searchr1, wang2025vrag-rl, liu2025visualarft, wu2025mmsearchr1, xiao2025m2io-r1}, where agents actively query textual and visual sources to retrieve evidence that augments their internal reasoning context. 
A complementary line is agentic visual processing~\cite{zheng2025deepeyes, cao2025ground, zhu2025active, zhang2025chain, su2025pixel, jiang2025vlm, su2025openthinkimg, zhang2025thyme, wu2025reinforcing, xu2025visual}, which equips agents with tools that transform, compose, or generate images during problem solving. 
For video understanding, recent tool-based methods usually interface with video through preprocessed surrogates: they convert videos into databases of dense captions or clip embeddings, design indexing schemes, and then let the agent query these static repositories with text-based tools~\cite{zhang2025DVD,tian2025ego-r1,long2025seeing}. 
In such systems, the agent can decide \emph{what} to ask but has little control over \emph{how} the video is observed; the temporal coverage, sampling density, and composition of segments are largely fixed by the preprocessing pipeline or a small set of coarse tool options.
\section{Method}
\label{sec:method}
\begin{figure*}[!tb]
    \centering
    % \vspace{-1em}
    \includegraphics[width=0.9\linewidth]{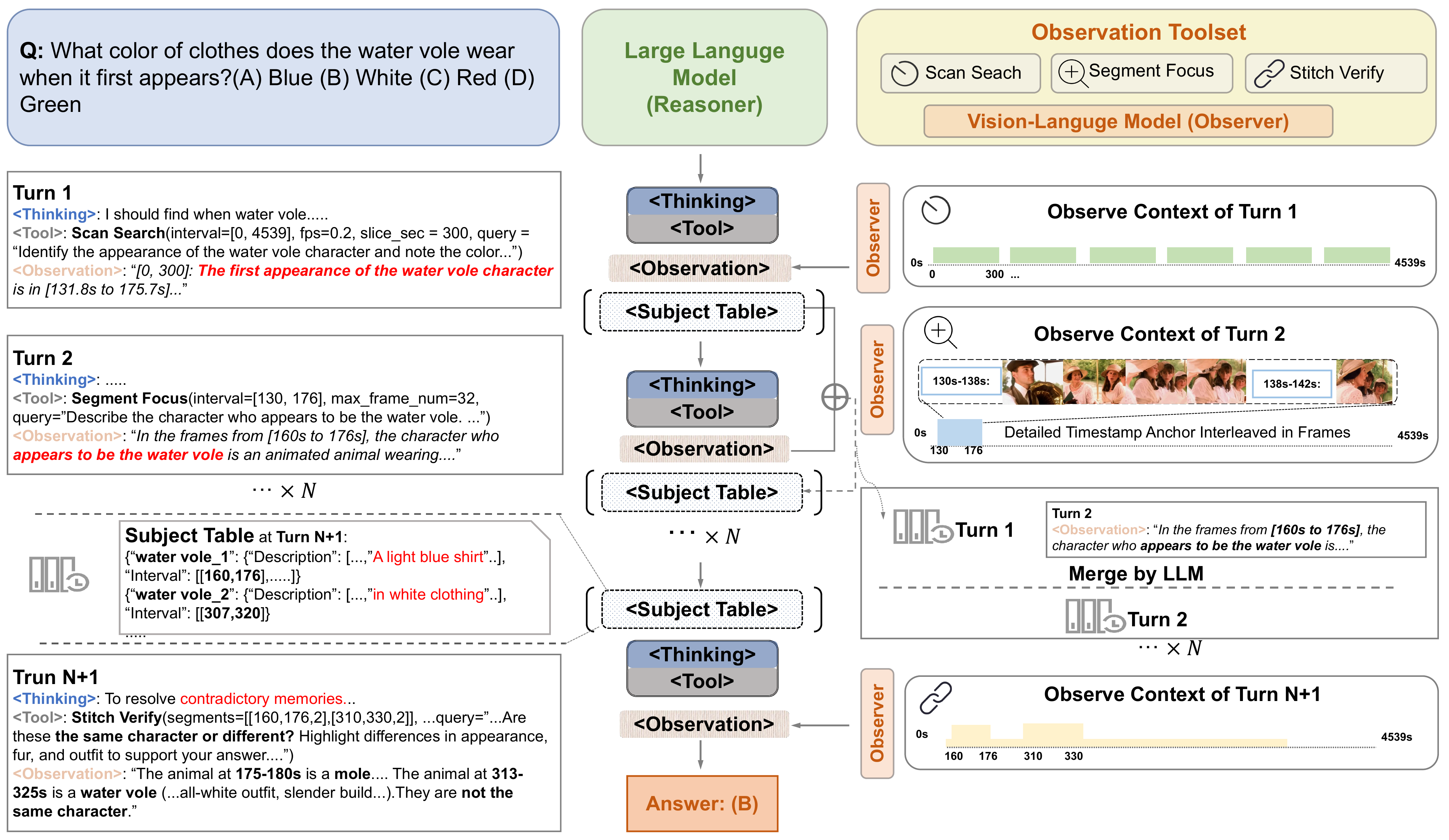}
    \vspace{-0.8em}
    \caption{Illustration of LensWalk’s reasoning-scheduled active observation on a real trace. The LLM reasoner alternates Scan Search and Segment Focus to seek question-relevant evidence, initially misclassifies a mole as the queried water vole and records conflicting memories in the subject table (red text), then invokes Stitch Verify to %jointly 
    inspect key segments, resolve the contradiction, and arrive at the correct answer.}
    \vspace{-1em}
    \label{fig:main_iilustartion}
\end{figure*}
% The challenge of video understanding lies in recovering high-level plot and causality from sparse visual signals spread over long durations. 
Despite recent great progress, most video understanding pipelines still pair strong reasoners with a fixed observation stage, so the scope and granularity of visual evidence do not adapt to the agent’s evolving thoughts. We instead view video understanding as also a problem of deciding how to observe, and LensWalk is designed precisely to let an agent dynamically choose where to look and at what granularity. see Appendix \ref{app:method-details} for pseudocode and concrete prompts.
\subsection{Overview of the LensWalk Architecture}
LensWalk models video understanding as an active observation process, driven by an agent that intelligently decides \textbf{where to look}, \textbf{what to observe}, and \textbf{how densely to sample} the video timeline per reasoning step. At its core, an LLM-based \textbf{Reasoner ($M_r$)} analyzes the user's query and accumulated evidence to formulate the next plan by calling observe tools with controllable video scopes. These tools belong to a versatile \textbf{Observation Toolkit ($\mathcal{O}$)}, which leverages a core VLM as its \textbf{Observer ($M_o$)} to extract information directly from a specific visible video context according to the Reasoner's arguments. To ensure coherence across multiple turns, \textit{Timestamp Anchors} and a global \textit{Subject Memory Table} are designed as lightweight memory mechanisms that provide evidence grounding. This establishes an iterative reason-plan-observe cycle. In the following subsections, we sequentially introduce how the agent schedules visual observations during its multi-turn tool use, the design of the observation toolkit, and the memory grounding mechanisms that ensure multi-turn coherence. Figure \ref{fig:main_iilustartion} illustrates the LensWalk agent pipeline on a real video understanding instance.
In the following subsections, we sequentially introduce how the agent schedules visual observations during its multi-turn tool use, the design of the observation toolkit, and the evidence grounding mechanisms that ensure multi-turn coherence.
% Table \ref{fig:lenswalk_pipeline} presents pseudocode outlining the LensWalk agent pipeline.
\subsection{Reason-Scheduled Video Observation}
To flexibly acquire evidence, the LensWalk agent schedules its visual observation by parameterizing each tool call based on its ongoing reasoning. At each step $t$, the \textbf{Reasoner} ($M_r$) integrates the user query $Q$, video metadata $V_{\text{info}}$ (e.g., duration, camera perspective), and the accumulated history of prior actions and observations $L_{t-1}$ to produce a fully specified plan, $a_t$.
$$
a_t = M_r(Q, V_{\text{info}}, L_{t-1}).
$$
This plan $a_t:=(o_t, q_t,\mathcal{I}_t,\rho_{o_t})$ is a tool call that precisely decides the next observation's scope and budget, as detailed below:
$$
% a_t := 
\left\{
\begin{array}{@{}l p{0.70\linewidth}@{}} % Using @{} removes column padding
o_t\in \mathcal{O} & The observation tool to invoke \\
q_t & The guiding sub-question \\
\mathcal{I}_t & Scope and Budget controls (e.g., temporal interval(s), frame count/FPS) \\
\rho_{o_t} & Tool-specific arguments 
\end{array}
\right.
$$
Executing this plan samples the specified frames (according to $\mathcal{I}_t$ and $\rho_{o_t}$) and, together with $q_t$, feeds them to the \textbf{Observer} ($M_o$), which returns evidence $e_t$ as a new observation. The plan-observation pair is then appended to the history as a new conversation turn: $L_t = [\,L_{t-1}, \{a_t, e_t\}\,].$ The explicit inclusion of controllable scope parameters $\mathcal{I}_t$ in each action is critical. It transforms the history $L_t$ into a structured document with precise temporal tags. Furthermore, this history of parameterized calls allows the agent to track its own exploration progress, enabling it to %identify and 
target under-explored video sections in subsequent steps.

\subsection{Multi-Granularity Observation Tool Suite}
% $\mathcal{O}$
The Reasoner’s plan is executed by a compact observation toolkit. Designed to enable diverse compositions of observation contexts, it offers a minimal yet expressive toolset: \textit{Scan Search} (broad localization), \textit{Segment Focus} (fine-grained inspection), and \textit{Stitched Verify} (cross-segment integration), plus a \textit{Finish} action. This effectively supports the reasoning-driven behaviors analyzed in Section~\ref{analysis on tool-call}. Figure \ref{fig:main_iilustartion} illustrates their arguments and examples.
% The Reasoner’s plan at is executed by a compact observation toolkit that acquires video evidence at complementary granularities. It offers three operators—\textit{Scan Search}, \textit{Segment Focus}, and \textit{Stitched Verify}—plus a Finish action for terminating reasoning once sufficient evidence is gathered; Figure \ref{fig:main_iilustartion} illustrates their arguments and examples.
\vspace{-1em}
% These tools form the sensory basis of the agent’s iterative reason–plan–observe loop: the Reasoner decomposes a complex query into sub-questions, selects the most suitable operator, and configures its parameters to progressively explore, refine, and verify visual hypotheses. By dynamically chaining these tools across rounds, the agent accumulates structured evidence directly from the raw video without precomputing a static representation or relying on external storage. 
% We introduce the three main observation tools in the following paragraphs, and Figure \ref{fig:main_iilustartion} illustrates their detailed arguments and examples.
\paragraph{Scan Search.}
This tool is designed for efficient, broad discovery across a specified temporal interval. In addition to a guiding sub-question ($q_t$) and scope controls ($\mathcal{I}_t$), the tool-specific settings ($\rho_{Scan}$) require a partitioning strategy, defined by either the number of slices ($n_s$) or the duration of each slice ($d_s$). The tool partitions the given interval, sparsely samples frames from each slice, and queries the VLM on a slice-by-slice basis. It returns per-slice summaries or relevance judgments that indicate the likely locations of target events. As each slice is processed independently, this operation is inherently parallelizable. This makes it well-suited for an initial exploration of still unfamiliar intervals or for detecting the presence of object or event cues, as its parallel nature allows for efficient localization of potentially relevant sections.
\vspace{-1em}
% This tool is designed for cheap, broad discovery over a global interval under the sample argument. The Reasoner supplies a time range to scan, together with either the number of slices or a slice duration, plus a low sampling rate and a guiding query. The tool partitions the interval, sparsely samples each slice, and queries the VLM slice by slice, returning per-slice summaries or relevance judgments that point to likely locations of target events. Because each slice is processed independently, the scan tool is naturally parallelizable and well suited for the first pass on long or information-sparse videos, or for “re-orienting” the agent when the current hypothesis lacks anchors. Its outputs—coarse timestamps and suggested sub-ranges—are ideal inputs for the more expensive tools below, enabling a direct scan $\rightarrow$ focus or scan $\rightarrow$ stitch handoff in the next turn. This contrasts with retrieval-based agents, where the exploration stage is mediated by a static clip store rather than the raw video timeline.  

\paragraph{Segment Focus.}
This tool executes a focused probe over a single, continuous time interval, applying a dense sampling strategy as defined by the scope controls ($\mathcal{I}_t$) to answer a precise sub-question ($q_t$). It is designed to extract fine-grained details, local dynamics, or small visual attributes that a coarser observation might miss. This operator is therefore ideal for hypothesis verification, attribute reading, or disambiguating similar actors and events. It also allows for re-examining the same interval with an even denser sampling configuration to clarify previously indistinct observations, a flexibility not easily available in methods reliant on pre-embedded video clips.
\vspace{-1em}
% Once a promising moment has been identified, the Reasoner can issue a focused probe over a single, tight interval with a higher fps or a stricter frame budget and a precise sub-question. This tool accepts one interval and global sampling settings; the Observer then prepares frames from that interval only and queries the VLM to extract detailed facts, local dynamics, or small visual attributes that a coarse scan would miss. Because the sampling budget is bounded per call, the Reasoner can raise density without jeopardizing the overall episode budget, which makes this operator a natural choice for hypothesis verification, attribute reading, or disambiguating similar actors/events flagged in the scan phase. Segment Focus is also the main vehicle for “I looked but it was blurry” situations: the agent can reissue the same interval with a denser configuration, something that is harder to express in agents whose observation is tied to pre-embedded clip stores.  

\paragraph{Stitched Verify.}
This tool is designed to observe facts that require integrating evidence from multiple, non-contiguous video segments, such as for before-and-after comparisons, tracking object transfers, or verifying causal narratives. Its tool-specific settings, $\rho_{Stitch}$, take a list of segments $S = [(s_i, e_i, r_i), \dots]$, where each entry defines a segment by its start time ($s_i$), end time ($e_i$), and a specific sampling rate for the segment ($r_i$). The operator consolidates frames from these distinct intervals into a single, coherent batch for the VLM. This uniquely supports asymmetric sampling—allocating a higher frame rate to action-heavy segments while using a lower rate for transitional ones. This enables the agent to verify a complex chain of events in a single step, reasoning over a unified visual context rather than reconciling isolated observations solely in the language domain.

\subsection{Evidence Grounding for Long-Horizon Coherence}
\label{sec:memory_and_grounding}
Because LensWalk runs multiple observation turns at different scopes, it must reconcile potentially ambiguous evidence across them. We therefore introduce two lightweight grounding components---\textit{Timestamp Anchors} and a \textit{Subject Memory Table}---that tie observations to precise temporal ranges and globally consistent entity references.
\vspace{-1em}
% This strategy enables the agent to integrate information into a coherent understanding and, when ambiguities persist, provides anchors to proactively plan subsequent observations to resolve them.
% as detailed in this subsection.
\paragraph{Timestamp Anchors within Tool Observation.}
During each observation, as in Figure \ref{fig:main_iilustartion}, detailed timestamp anchors at a specified granularity would be inserted to \textit{Observer} its visual context(frame sequences). These anchors are inserted directly within the tool's visible frames as text, prompting the \textit{Observer} to ground its answers with explicit temporal references (e.g., “at 01:15-01:40”). This provides the \textit{Reasoner} with fine-grained temporal cues, enabling it to efficiently pinpoint evidence sources in subsequent planning cycles, schedule re-observations over ambiguous ranges, and prune irrelevant sections for more precise analysis.
\vspace{-1em}
\paragraph{Subject Memory Table for Entity Consistency.}
After each observation, as in Figure \ref{fig:main_iilustartion}, a global \textit{Subject Memory Table} $Sub_{t}$ is updated by an LLM with entity information from the new tool observation. It maintains outside the main reasoning history $L_t$, records persistent entities—such as people or objects—along with summarized attributes and seen timestamps. Before each reasoning turn, it would be inserted into the context and overwrite the previous version. This evolving memory serves a dual purpose: it allows the Reasoner to use canonical identifiers for efficient, consistent entity reference, eliminating the need to repeatedly re-identify entities from the tool history in each reasoning cycle. It also provides a clean, structured knowledge base that the Reasoner consults to strategically plan its subsequent observations. This mechanism manages accumulating entity information to maintain cross-round reference stability, and avoids redundant context tokens from lengthy descriptions, ensuring the understanding of subjects remains consistent as it progressively explores the video.

Maintaining consistent, reliable memory across multiple reasoning turns is challenging in agentic video understanding~\cite{pang2025mrvideo,zhang2025DVD,long2025seeing}, and prior systems often resort to heavy pre-scanning, summarization, or intricate bookkeeping schemes~\cite{tian2025ego-r1,long2025seeing,wang2024videoagent,zhang2025DVD}, which incur considerable computational and engineering overhead. Our evidence grounding instead maintains a compact, structured state and, by tying evidence to temporal scopes and entities in the history, allows the agent to revisit ambiguous regions via re-observation, as illustrated in Figure~\ref{fig:main_iilustartion}. Although Observer or memory-updating errors can still introduce inconsistencies, this lightweight combination of grounding and re-observation yields a scalable, self-corrective pipeline that fits into the reason--plan--observe loop and is supported by the ablations in Section~\ref{subsec:ablation}.

% Maintaining a consistent and reliable memory across multiple reasoning turns has been a significant challenge in agentic video understanding\cite{pang2025mrvideo,zhang2025DVD,long2025seeing}. Previous works have often resorted to either exhaustive pre-scanning and extensive summarization to eliminate ambiguity\cite{tian2025ego-r1,long2025seeing}, or have relied on elaborately designed cross-validation rules and auxiliary data structures to preserve all potential memories\cite{wang2024videoagent,zhang2025DVD,pang2025mrvideo}. Both approaches demand considerable upfront computational or design effort.

% Ambiguity is inherent in complex video content, even for human viewers. While our Evidence Grounding components cannot completely resolve all ambiguities and contradictions encountered, owing to potential errors from the \textit{Observer} or the memory-updating model, a key advantage of our framework is the agent's ability to resolve such inconsistencies by actively planning new observations. Because the agent can freely schedule observations and trace evidence back to its source video scope via the history, it can target these ambiguities with new, focused tool calls to investigate and resolve them, as illustrated in Figure \ref{fig:main_iilustartion}. The empirical effectiveness of these strategies is also demonstrated in our ablation study (Section \ref{subsec:ablation}). Therefore, our framework provides a scalable working memory that dynamically evolves and self-corrects through the continuous interplay of reasoning and observation

\section{Experiments}
\label{sec:exp}
\begin{table*}[t]
    \caption{
    \centering
    Comparison on long video benchmarks. For LensWalk, “(·/·)” denotes Rasoner/Observer; “($M$)” means “($M$/$M$)”.
    % \\  \textit{Note: In “LensWalk (X / Y)”, X denotes the reasoner and Y denotes the observer.}
    }
\label{tab:longvideo}

    \label{tab:longvideoresults}
    % \vspace{-2mm}
    \centering
    \renewcommand\arraystretch{1.0} % Adjust this value to control row height
    {
    \small
    \setlength{\tabcolsep}{2pt}        % 只在此表更小
    \renewcommand{\arraystretch}{0.88} % 只在此表更小
    \setlength{\aboverulesep}{0.2ex}
    \setlength{\belowrulesep}{0.2ex}
    % \resizebox{0.79\linewidth}{!}
    {
        \begin{tabular}{l cccc}
            \toprule
            \multirow{2}{*}{%
  \begin{tabular}[c]{@{}l@{}}%
    \textbf{Methods}\\[-0.2ex]
    \textit{(Reasoner/Observer)}%
  \end{tabular}%
}   & \textbf{LVBench} & \textbf{LongVideoBench}    & \textbf{Video MME}    & \textbf{EgoSchema} \\
            ~       & Overall & (Val) Long  & Long (w/o sub)  & Val \\
            \midrule
            \multicolumn{4}{l}{\small\textcolor{gray}{\textit{Commercial VLMs}}} \\
            Gemini-1.5-Pro~\cite{team2024gemini1.5}  & 33.1 & 58.6 & 67.4  & -  \\
            Gemini-2.0-Flash~\cite{GoogleDeepMind2024Gemini2}  & 48.3 & 45.7 & 63.0  & 71.2  \\
            GPT-4o~\cite{hurst2024gpt} & 48.9  & 60.9 & 65.3  & 70.4  \\
            GPT-4.1~\cite{openai2025gpt4.1} & 51.9  & 64.6 & 63.1  & 72.2  \\
            o3~\cite{OpenAI2025o3o4mini} & 57.1  & 60.6 & 64.7 & 63.2 \\
            GPT-5~\cite{openai2025gpt5} & 59.8  & 61.8 & 68.4 & 73.8 \\
            \midrule
            \multicolumn{4}{l}{\small\textcolor{gray}{\textit{Open-Source VLMs}}} \\
            % TimeMarker & 41.3  & -  & 46.4  & - \\
            LLaVA-Video-72B~\cite{zhang2024video} & 38.7  & - & 59.6  & 74.7  \\
            InternVL2.5-78B~\cite{chen2024internvl2.5} & 43.6  & - & 62.6 & -  \\
            Qwen2.5-VL-72B~\cite{bai2025qwen2} & 47.7  & 54.2 & 63.1  & 75.4  \\
            AdaReTaKe~\cite{wang2025adaretake} & 53.3  & - & 65.0 & -  \\
            \midrule
            \multicolumn{4}{l}{\small\textcolor{gray}{\textit{Video Agents and Others}}} \\
            % VideoTree & 28.8 & - & -  & 67.0  \\
            VideoAgent~\cite{wang2024videoagent} & 29.3  & - & - & 63.2  \\
            VCA~\cite{yang2025vca} & 41.3 & - & -  & 73.6  \\
            Ego-R1~\cite{tian2025ego-r1} & - & - & 64.9 & 68.2 \\
            MR. Video~\cite{pang2025mrvideo} & 60.8 & 61.6 & 61.8 & 73.0  \\
            Deep Video Discovery~\cite{zhang2025DVD} & \textbf{74.2} & 68.6  & 67.3 & 76.6  \\
            \midrule
            \textbf{LensWalk (o3/GPT-4.1)} & 66.8 & 69.9 & 70.0 & 74.8\\
            \textbf{LensWalk (o3/Qwen2.5-VL-72B)} & 59.4 &67.3 & 66.7 & \textbf{77.2}\\
            \textbf{LensWalk (o3)} &68.6 & \textbf{70.6} & \textbf{71.4} &74.8 \\
            \textbf{LensWalk (GPT-5)} &66.9 &68.8 & 69.2 & 74.6 \\
            \bottomrule
        \end{tabular}
    }
    }
    % \label{tab:long_video_benchmarks}
    \vspace{-3pt}
\end{table*}
\subsection{Setting}

\paragraph{Datasets.} %Following prior work, 
We conduct experiments on challenging long-video understanding benchmarks. Specifically, we evaluate LensWalk on LVBench~\cite{wang2025lvbench} and LongVideoBench (long split of validation set)~\cite{wu2024longvideobench} for extended temporal comprehension, and on 30\-60min split (long split) of Video-MME~\cite{fu2025videomme} for general long-video understanding. In addition, we report results on two well-established video reasoning benchmarks, MMVU~\cite{zhao2025mmvu} and Video-MMMU~\cite{hu2025video}, to assess how LensWalk helps on video QA tasks that require higher-level reasoning. We also include EgoSchema~\cite{mangalam2023egoschema} to cover egocentric video understanding.

\paragraph{Baselines.} We compare against a broad set of methods for video understanding: (1) VLM-based, including both proprietary and open-source vision-language models such as o3~\cite{OpenAI2025o3o4mini}, GPT‑4.1~\cite{openai2025gpt4.1}, Gemini-2.5 Pro~\cite{GoogleDeepMind2025Gemini25}, and Qwen2.5-VL-72B~\cite{bai2025qwen2}. (3) Agentic Video Understanding methods, such as VideoAgent~\cite{wang2024videoagent}, VCA~\cite{yang2025vca}, MR.Video~\cite{pang2025mrvideo} and Deep Video Discovery~\cite{zhang2025DVD}. Most reported results are from official leaderboards or published reports, while we reproduce the results of GPT-4.1 and GPT-5~\cite{openai2025gpt5} via official API, with uniformly sampling 256 frames following~\cite{zhang2025DVD, pang2025mrvideo}. More detailed configuration of baselines is in Appendix \ref{sec:Evaluation Details}.
\paragraph{Implementation Details.}
As a plug-and-play framework, LensWalk accommodates any combination of reasoning-based language and vision-language models. To demonstrate its potential, we choose combinations of state-of-the-art models such as Qwen2.5-VL-72B, GPT-4.1, o3, and GPT-5(with minimal reasoning effort, both in baseline and ours), with the Reasoner model also serving as the memory updater by default. Proprietary models are queried via their official APIs. We constrain the agent to a maximum of 20 tool invocations, one per turn. The per-call frame budgets for Scan Search, Segment Focus, and Stitched Verify are set at 180, 32, and 128, respectively, allowing the agent to allocate its observation budget freely within these limits.

\begin{table*}[t]
    \begin{minipage}[t]{0.6\textwidth} % 调整了宽度以适应新列
    \caption{
    Comparison on video reasoning benchmarks. For LensWalk, “(·/·)” \\denotes Reasoner/Observer; “($M$)” means “($M$/$M$)”.
    % {In “LensWalk (X / Y)”, X denotes the reasoner and Y denotes the observer.}
    }
    \label{tab:reason_benchmarks}
    \centering
    \renewcommand\arraystretch{1.15}
    {
    % \resizebox{\linewidth}{!}
    \small
        \setlength{\tabcolsep}{3pt}        
        \renewcommand{\arraystretch}{0.88} 
        \setlength{\aboverulesep}{0.2ex}
        \setlength{\belowrulesep}{0.2ex}
    {
    \begin{tabular}{l c cccc}
            \toprule
            \multirow{2}{*}{%
  \begin{tabular}[c]{@{}l@{}}%
    \textbf{Methods}\\[-0.2ex]
    \textit{(Reasoner/Observer)}%
  \end{tabular}%
} & \textbf{MMVU} & \multicolumn{4}{c}{\textbf{Video-MMMU}} \\
            \cmidrule(lr){2-2} \cmidrule(lr){3-5} \cmidrule(lr){6-6}
           ~ & MC & Percep. & Compreh. & Adap. & Overall \\
           % ~ & MC & Per & Comp& Ada & Overall \\
\midrule
Gemini-1.5-Flash\cite{team2024gemini1.5} & 66.6 & 57.33 & 49.00 & 43.00 & 49.78 \\
Gemini-1.5-Pro\cite{team2024gemini1.5} &74.7 & 59.00 & 53.33 & 49.33 & 53.89 \\
GPT-4o\cite{hurst2024gpt}& 78.1 & 66.00 & 62.00 & 55.67 & 61.22 \\
GPT-4.1\cite{openai2025gpt4.1} & 76.3 & 76.00 & 70.67 & 55.67 & 67.44 \\
o3\cite{OpenAI2025o3o4mini} & 78.9 & 79.33 & 75.67 & 71.33 & 75.44\\
\midrule
Aria\cite{li2024aria} & 60.6& 65.67 & 46.67 & 40.00 & 50.78 \\
InternVL2-8B\cite{chen2024internvl2} & 49.1 & 47.33 & 33.33 & 31.67 & 37.44 \\
Qwen2.5-VL-7B\cite{bai2025qwen2} & 60.6 & 58.33 & 44.33 & 39.67 & 47.44 \\
Qwen-2.5-VL-72B\cite{bai2025qwen2} &69.3 & 69.33 & 61.00 & 50.33 & 60.22 \\
% LLaVA-Video-72B & & 59.67 & 46.00 & 43.33 & 49.67 \\
\midrule
\textbf{LensWalk (o3/GPT-4.1)} & \textbf{80.9} & \textbf{82.00} & 77.67 & 71.67 & 77.11\\
\textbf{LensWalk (o3)} & 79.2 & 81.33 & \textbf{81.33} & \textbf{72.33} & \textbf{78.33}\\
            \bottomrule
        \end{tabular}
    }
    }
    % \label{tab:ablation-tools-extended}
    % \vspace{-5mm}
\end{minipage}
    \hfill
    \begin{minipage}[t]{0.38\linewidth}
        \caption{Results on more Reasoner and Observer choices on VideoMME (long split).}
        \label{tab:opensourcellm}
        \centering
        \renewcommand\arraystretch{1.15}
        {
        % \resizebox{\linewidth}{!}
        \small
        \setlength{\tabcolsep}{3pt}        
        \renewcommand{\arraystretch}{0.88} 
        \setlength{\aboverulesep}{0.2ex}
        \setlength{\belowrulesep}{0.2ex}
        {
        \begin{tabular}{llc}
    \toprule
    \textbf{Observer} & \textbf{Reasoner} & \textbf{Acc. (\%)} \\
    \midrule
    \multirow{3}{*}{GPT-4.1}
        & -             & 63.1 \\
        & o3                 & 70.0 (\textcolor{green}{+6.9}) \\
        & {\scriptsize Qwen3-235B-A22B}    & 63.2
        % (?check为啥两段差这么多) 
        (\textcolor{green}{+0.1}) \\
    \midrule
    \multirow{3}{*}{GPT-4.1-mini}
        & -             & 59.4 \\
        & o3                 & 66.0 (\textcolor{green}{+6.6}) \\
        & {\scriptsize Qwen3-235B-A22B}   & 61.0 (\textcolor{green}{+1.6})\\
    \midrule
    \multirow{3}{*}{Qwen2.5-VL-72B}
        & -             & 63.1 \\
        & o3                 & 66.7 (\textcolor{green}{+3.6}) \\
        & {\scriptsize Qwen3-235B-A22B}    & 62.5 (\textcolor{red}{-0.6})\\
    \midrule
    \multirow{3}{*}{Qwen2.5-VL-7B}
        & -             & 55.4 \\
        & o3                 & 61.3 (\textcolor{green}{+5.9}) \\
        & {\scriptsize Qwen3-235B-A22B}    & 59.7 (\textcolor{green}{+4.3}) \\
    \bottomrule
\end{tabular}
        }
        }
    \end{minipage}
    \vspace{-5pt} 
\end{table*}

\subsection{Main Results}
As demonstrated in Table \ref{tab:longvideoresults} and \ref{tab:reason_benchmarks}, LensWalk achieves strong performance across long video understanding and video reasoning benchmarks. The framework significantly enhances powerful base models, outperforming all \textit{Observer} baselines and multimodal \textit{Reasoners} like o3 and GPT-5 that reason directly over video frames. LensWalk consistently surpasses prior agentic methods, notably achieving over 4\% higher accuracy on the challenging VideoMME-Long benchmark than leading approaches like Mr. Video and Deep Video Discovery, which necessitate a vast number of extra captions to support their reasoning. On the reasoning-focused MMVU and Video-MMMU benchmarks, our framework's ability to schedule observations for the specialized o3 model boosts accuracy by an additional 2\% and 3\%. Beyond absolute gains, LensWalk also enables complementary pairing: using o3 as Reasoner over GPT-4.1 observations surpasses either model alone on MMVU, and a “self-play” configuration where a single multimodal reasoning model like o3 serves as both Reasoner and Observer provides strong plug-and-play improvements without any fine-tuning. We report additional shared-model comparisons of other agents and a static extracted-frames baseline in Appendix Table~\ref{tab:shared_o3_lvbench}, \ref{tab:additional_reasoners_static}.

% Furthermore, an intriguing synergy emerges within the framework. Pairing the stronger o3 Reasoner with observations from GPT-4.1 allows it to surpass its own performance when observing the video directly, highlighting the framework's capacity to pool complementary model strengths. A "self-play" approach also proves compelling, yielding significant gains when a single multimodal model serves as both Reasoner and Observer. This "free lunch" underscores a symbiotic relationship for complex long-video tasks: progressive reasoning ensures essential evidence is not overlooked, while targeted observation provides the context to forge more effective reasoning paths.

\begin{table}[!t] % 也可用 [h] [b]
  \centering
  \caption{\small Ablation on the Observation Tools and Reasoning coherence modules of ours(o3/GPT-4.1) on Video-MME (long split).}
  \label{tab:ablation-tools-extended}
  \setlength{\tabcolsep}{5pt} % 调整列间距
  \renewcommand{\arraystretch}{1.2} % 调整行高
  {
  \small
  \setlength{\tabcolsep}{2pt}        
  \renewcommand{\arraystretch}{0.88} 
  \setlength{\aboverulesep}{0.2ex}
  \setlength{\belowrulesep}{0.2ex}
  \begin{tabular}{ccc|cc|c}
        \toprule
        \multicolumn{3}{c|}{\textbf{Observation Tools}} & \multicolumn{2}{c|}{\textbf{Reasoning Coherence}} & \multirow{2}{*}[-1ex]{\makecell{VideoMME \\ Long}} \\ 
        \cmidrule(lr){1-5}
        \makecell{Scan\\Search} & 
        \makecell{Segment\\Focus} & 
        \makecell{Stitch\\Verify} &
        \makecell{Timestamp\\Anchor} &
        \makecell{Subject\\Memory} & \\  % <--- 关键修改：为multirow跨越的行留出空位
        \midrule
                    & \checkmark    & \checkmark    & & &  65.4  \\
        \checkmark  &               & \checkmark    & & &  68.1 \\
        \checkmark  & \checkmark    &               & & &  66.8 \\
        \midrule
        \checkmark  & \checkmark    & \checkmark    & \checkmark & & 69.7 \\
        \checkmark  & \checkmark    & \checkmark    & & \checkmark & 69.4 \\
        \midrule
        \textbf{\checkmark}  & \checkmark    & \checkmark & \checkmark & \checkmark &  \textbf{70.0} \\
        \bottomrule
        \end{tabular}
  }
\vspace{-1em}
\end{table}

\subsection{Ablation and Analysis}
\label{subsec:ablation}
\paragraph{Applicability to Open-Source Models.} As shown in Table \ref{tab:opensourcellm}, we validated the framework's generality by substituting various models as \textit{Reasoner} and \textit{Observer}. The results indicate that LensWalk effectively boosts weaker \textit{Observers}, such as Qwen2.5-VL-7B. Moreover, the open-source \textit{Reasoner} Qwen3-235B-A22B~\cite{yang2025qwen3} demonstrates a positive impact, notably improving Qwen2.5-VL-7B by 4.3\%. However, we observed this open-source \textit{Reasoner} struggles to enhance stronger \textit{Observers} like GPT-4.1 (+0.1\%) and Qwen2.5-VL-72B (-0.6\%), suggesting its generated observation plans are less effective than these models' innate reasoning capabilities. This finding highlights the decisive role of the \textit{Reasoner}'s cognitive strength in guiding the acquisition of crucial visual evidence. See more results on more open-source reasoner models in Appendix Table~\ref{tab:additional_reasoners_static}
\paragraph{Ablation on Components.} We conducted an ablation study to assess the contribution of each component on VideoMME(long split), with results shown in Table \ref{tab:ablation-tools-extended}. The observation tools prove to be highly complementary. Removing \textit{Scan Search}, which is critical for efficiently discovering cues, leads to the largest accuracy drop of 4.6\%. The absence of \textit{Stitch Verify}, which impacts the ability to connect disparate scenes for causal analysis, and \textit{Segment Focus}, for extracting fine-grained facts, also degrades performance by 3.2\% and 1.9\%, respectively. Furthermore, the reasoning coherence modules provide distinct benefits. Adding the \textit{Timestamp Anchor} boosts performance to 69.7\%, and incorporating the \textit{Subject Memory} raises it to the final 70.0\%, underscoring the importance of both fine-grained temporal alignment and consistent entity tracking in video observation-reason loops.
% 对应两个表
% 对各个工具和时间戳、记忆角色表两个模块的性能消融
% 对不同的reasoner observer

\begin{figure}[!tb]
    \centering
    % \vspace{-1em}
    \includegraphics[width=0.9\linewidth]{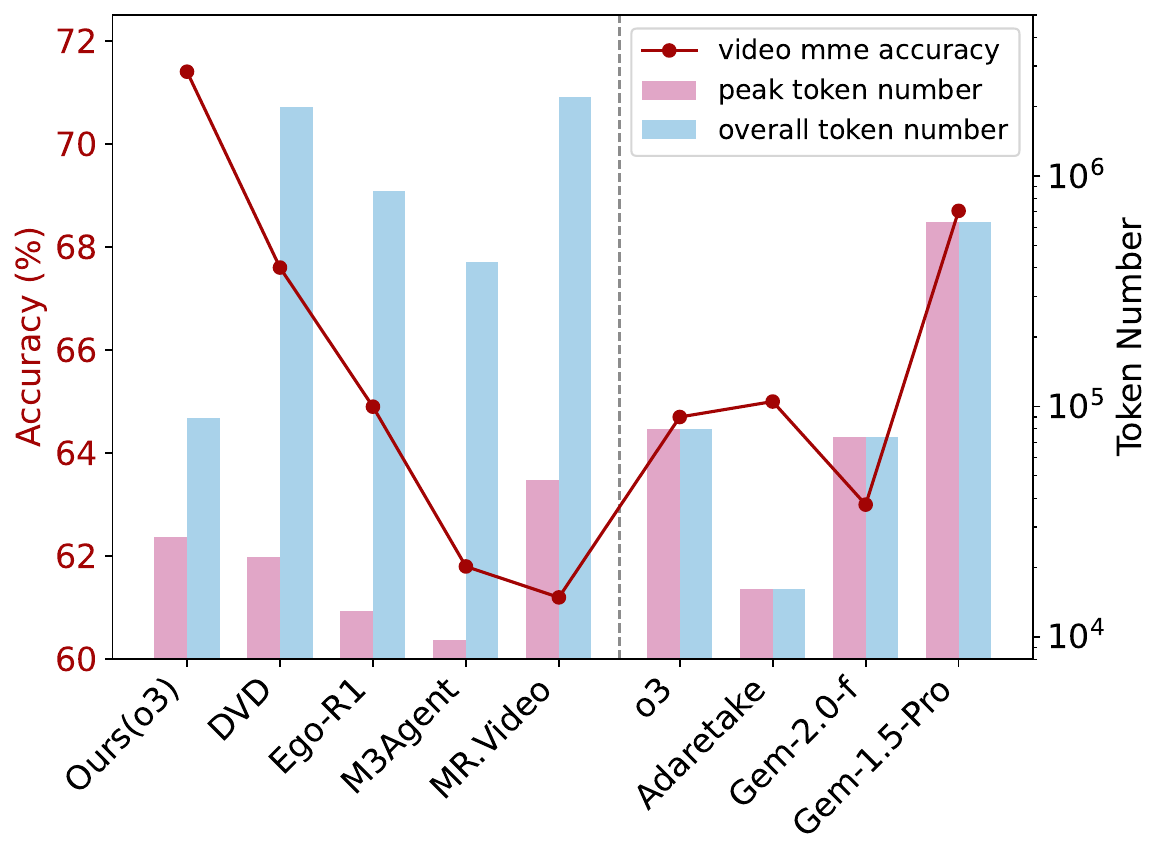}
    \vspace{-1.6em}
    \caption{Comparison of video understanding methods on accuracy and token efficiency. Abbrev. of \textit{Gemini}: Gem.}
    \vspace{-1.5em}
    \label{fig:acc_and_cost}
    \vspace{-3pt} 
\end{figure}

\paragraph{Efficiency Analysis.} 
To provide a fair comparison of efficiency independent of model deployment or network conditions, we mainly use two key metrics: total token cost and peak per-turn context tokens. As in Figure \ref{fig:acc_and_cost}, our reasoning-driven framework avoids the substantial upfront cost of prior methods like Deep Video Discovery (DVD) and Ego-R1 that rely on pre-generating captions for an entire video, much of which is irrelevant. Further comparisons in Appendix (Table \ref{tab:shared_o3_lvbench}) show that LensWalk achieves competitive accuracy with substantially fewer frames and much less end-to-end inference time.
Furthermore, while consuming total tokens comparable to single-forward baselines, our framework's reasoning-driven schedule strategically observes across multiple turns. This approach not only enhances performance but also significantly lowers the peak token count per turn, alleviating the severe memory pressure and throughput bottlenecks of processing a massive, monolithic context. In addition, as detailed in Appendix (Table \ref{tab:adaptive_budget_scaling}), the framework adaptively allocates its observation budget, converging quickly on simple queries while scaling computation for longer or reasoning-intensive cases.
% In summary, LensWalk achieves efficiency by distributing visual analysis across guided steps, avoiding both the exhaustive computation of full-video captioning and the peak memory demands of single-pass methods.
% To provide a fair comparison independent of model deployment, hardware, or network conditions, we evaluate LensWalk's efficiency using two key metrics: total token cost and peak per-turn context tokens. As in Figure \ref{fig:acc_and_cost}, our reasoning-driven framework avoids the substantial upfront cost of prior methods like Deep Video Discovery (DVD) and Ego-R1 that rely on pre-generating captions for an entire video, much of which is irrelevant. Furthermore, while the total tokens processed are comparable to or slightly beyond single-forward baselines, our framework's reasoning-driven schedule strategically observes across multiple turns. This approach not only enhances performance but also significantly lowers the peak token count per turn, alleviating the severe memory pressure and throughput bottlenecks of processing a massive, monolithic context. In summary, LensWalk achieves efficiency by distributing visual analysis across guided steps, avoiding both the exhaustive computation of full-video captioning and the peak memory demands of single-pass methods.
\vspace{-3pt} 
\begin{figure*}[tb]
    \centering
    % \vspace{-1em}
    \includegraphics[width=0.9\linewidth]{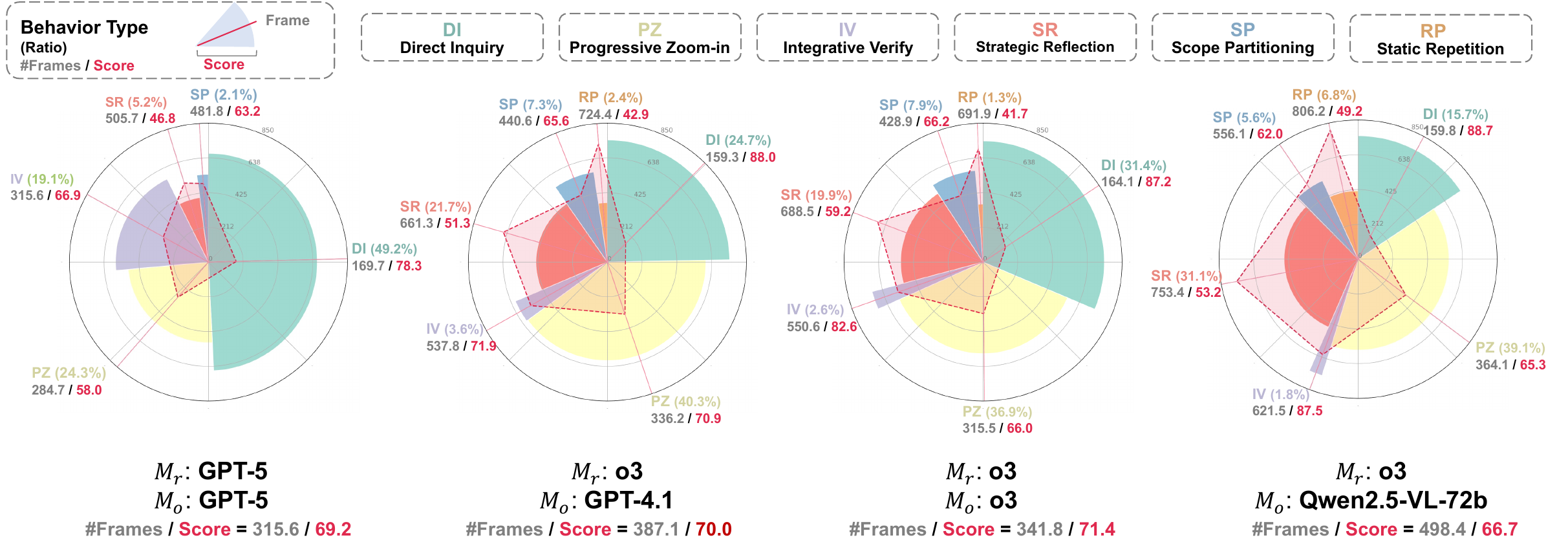}
    %shenhongze 换字体后导出后换成正式pdf
    % \vspace{-1.6em}
    \caption{Tool-call behavior of LensWalk across model recipes, visualized following~\cite{zhang2025deep}. We group behaviour into six strategy types and show their ratio (sector angle), average frame cost (dashed polygon), and accuracy (noted as score; sector radius), revealing adaptive allocation of more frames to harder queries and diverse exploration under uncertainty. }
    \vspace{-1em}
    \label{fig:trace_type}
\end{figure*}
\subsection{Analysis on Tool-Call Behaviour}
\label{analysis on tool-call}
Our design empowers the agent to control its observation scope, enabling it to devise diverse observational paths based on accumulated experience. To better analyze this, we examine the agent's tool calls and identify distinct behavioral patterns by considering the observation query, the specific tool selected, and the observational scope at each step. We categorize these patterns into six primary modes (in Figure \ref{fig:trace_type}, see Appendix \ref{app:more results} for details and examples.):\\
% \textbf{Direct Inquiry}: Resolving a task in a single, confident step without prior exploration, typically for high-level or temporally-anchored questions. \\
% \textbf{Progressive Zoom-in}: Progressively narrowing the search scope, starting with a broad scan to identify key moments and then zooming into these promising segments for a more detailed inspection. \\
% \textbf{Scope Partitioning}: Partitioning the timeline to analyze distinct scopes independently, then integrating the resulting insights purely at a reasoning level to form a conclusion. \\
% \textbf{Strategic Reflection}: After a focused search hits a dead end, strategically pausing to reflect and re-evaluate the approach. This leads to a new, broad scan to find alternative leads or different regions of interest. \\
% \textbf{Integrative Verify}: Verifying a complex hypothesis by integrating the observation scopes from multiple, distinct key moments. This holistic observation confirms the conclusion. \\
% \textbf{Static Repetition}: Falling into an unproductive cycle of repetitive observations. This involves examining the same region multiple times with a static intent, showing no progress or change in strategy. 
\textbf{Direct Inquiry.} Resolving a task in a single, confident step without prior exploration, typically for high-level or temporally anchored questions.

\textbf{Progressive Zoom-in.} Starting from a broad scan to locate candidate moments and then zooming into promising segments for more detailed inspection.

\textbf{Scope Partitioning.} Partitioning the timeline into distinct scopes, analyzing them independently, and integrating the resulting insights purely at the reasoning level to form a conclusion.

\textbf{Strategic Reflection.} After a focused search hits a dead end, pausing to reassess the approach and launching a new, broader scan over alternative regions or hypotheses.

\textbf{Integrative Verify.} Verifying a complex hypothesis by jointly inspecting multiple key moments in a single stitched observation, enabling holistic confirmation of the conclusion.

\textbf{Static Repetition.} Falling into an unproductive cycle of nearly identical observations over the same region with a static intent, showing little progress or change in strategy.

Analysis of these patterns reveals several key findings:
\begin{itemize}
    % \item \textbf{The agent demonstrates difficulty-aware cost allocation.} It intuitively allocates more observational effort to challenging problems. A clear trend shows that effort is inversely proportional to success: long, costly observation traces are spent on difficult tasks, while simple cases are resolved efficiently with minimal inquiry. This indicates the agent intelligently adapts its strategy based on its unfolding observations, deciding whether a deeper, more expensive investigation is warranted.
    \item \textbf{Difficulty-aware cost allocation.} The agent intuitively allocates more observational effort to challenging problems. Hard questions with lower accuracy are associated with longer, more costly observation traces, while simple cases are often resolved efficiently with less frames or even single-step inquiries. This indicates that LensWalk adapts its observation depth based on the evolving uncertainty of each query.
    
    % \item \textbf{Emergent reflective strategies replace inefficient loops.} Empowered by diverse choices in its observation plan, our agent effectively avoids most of 'Static Repetition' common in prior work. When faced with uncertainty, it substitutes these failure modes with reflective behaviours like \textit{Strategic Reflection} or \textit{Integrative Verify} to expand its evidence and consolidate information. As powerful recovery mechanisms, they yield far more stable and higher accuracy than the rare looping failures observed. This demonstrates that granting an agent flexible visual scheduling capabilities allows it to develop more resilient, human-like problem-solving strategies.
     \item \textbf{Emergent reflective strategies replace inefficient loops.} Empowered by diverse choices in its observation plan, the agent largely avoids the ``Static Repetition'' behaviour reported in prior work\cite{zhang2025DVD,tian2025ego-r1}. When faced with uncertainty, it instead invokes reflective patterns such as \emph{Strategic Reflection} or \emph{Integrative Verify} to expand and consolidate evidence. These recovery mechanisms yield more stable and higher accuracy than the rare looping failures that remain.
    % \item  \textbf{Observer quality dictates reasoning strategy and efficiency.} For a fixed reasoning agent, the observer's capability is the critical performance driver. A strong observer (e.g., o3) delivers clear, decisive evidence that directly addresses the query's intent, enabling efficient, low-cost resolutions. Conversely, a weaker observer with weaker perception and instruction-following (e.g., Qwen2.5-VL-72B vs o3) introduces ambiguity, forcing the agent into complex and expensive observation plans like retreating or verifying. This efficiency gap is profound: even when correct(e.g. Direct Inquiry), the agent requires a more costly investigation with a weaker observer, proving high-quality perception is paramount to effective reasoning.
     \item \textbf{Observer quality shapes strategy and efficiency.} For a fixed Reasoner, the Observer’s capability is a critical performance driver. A strong Observer (e.g., o3) supplies clear, decisive evidence that directly addresses the query’s intent, enabling frequent, low-cost \emph{Direct Inquiry}. In contrast, weaker Observers (e.g., Qwen2.5-VL-72B) introduce ambiguity, pushing the agent into more complex and expensive observation plans such as retreating to broader scans or repeated verification. This gap highlights that high-quality perception is paramount for cost-effective reasoning.
\end{itemize}
% \paragraph{Qualitative Comparison}
\section{Conclusion}
In this work, we introduced LensWalk, an agentic framework that actively scheduled video observation based on reasoning in each iteration. By granting a language model the ability to dynamically control what to see, and at what granularity, LensWalk transforms video understanding from a static recognition task into an adaptive reason-plan-observe loop. This design not only yields substantial, plug-and-play performance gains across diverse multimodal models and long-video benchmarks, but also uncovers emergent cognitive behaviors such as progressive zoom-in, reflective replanning, and integrative verification. 
Our findings highlight that scalable video intelligence lies not only in ever-larger frozen representations, but in scalable visual cognition—where observation is actively orchestrated by reasoning intent. We believe LensWalk provides a concrete step towards the next generation of self-directed multimodal agents, capable of learning how to observe as they think, and ultimately, how to think through seeing.

%In this work, we introduced LensWalk, an agentic framework that transforms video understanding from a static recognition pipeline into an adaptive reason–plan–observe loop. By allowing a language model to dynamically determine what to observe and at what granularity, LensWalk enables the progressive acquisition of targeted visual evidence and supports on-demand refinement of evolving hypotheses. This design yields substantial, plug-and-play performance gains across diverse multimodal models and long-video benchmarks, while also giving rise to emergent behaviors—such as multi-hop verification, reflective replanning, and trial-and-error recovery—that reflect tighter coupling between perception and reasoning.

%These results highlight a broader perspective: effective long-video understanding depends not only on stronger encoders or larger context windows, but on an agent’s ability to manage its own perception in response to uncertainty and evolving goals. By reframing observation as an active decision process, LensWalk provides a foundation for future multimodal agents that reason more reliably, allocate computation more efficiently, and ultimately improve their reasoning by deciding how and where to look.
\section*{Acknowlegment}
This work is supported by the National Natural Science Foundation of China (Nos.62495084, 62376259).
{
    \small
    \bibliographystyle{ieeenat_fullname}
    \bibliography{main}

@article{chen2024mindsearch,
  title   = {{MindSearch}: Mimicking human minds elicits deep {AI} searcher},
  author  = {Chen, Zehui and Liu, Kuikun and Wang, Qiuchen and Liu, Jiangning and Zhang, Wenwei and Chen, Kai and Zhao, Feng},
  journal = {arXiv},
  year    = {2024},
  eprint  = {2407.20183},
  eprinttype = {arXiv},
  eprintclass = {cs.AI},
  note    = {arXiv:2407.20183}
}

@article{zhang2025DVD,
  title   = {{DVD}: Deep video discovery: Agentic search with tool use for long-form video understanding},
  author  = {Zhang, Xiaoyi and Jia, Zhaoyang and Guo, Zongyu and Li, Jiahao and Li, Bin and Li, Houqiang and Lu, Yan},
  journal = {arXiv},
  year    = {2025},
  eprint  = {2505.18079},
  eprinttype = {arXiv},
  eprintclass = {cs.CV},
  note    = {arXiv:2505.18079}
}

@article{feng2025retool,
  title   = {{ReTool}: Reinforcement learning for strategic tool use in {LLM}s},
  author  = {Feng, Jiazhan and Huang, Shijue and Qu, Xingwei and Zhang, Ge and Qin, Yujia and Zhong, Baoquan and Jiang, Chengquan and Chi, Jinxin and Zhong, Wanjun},
  journal = {arXiv},
  year    = {2025},
  eprint  = {2504.11536},
  eprinttype = {arXiv},
  eprintclass = {cs.LG},
  note    = {arXiv:2504.11536}
}

@article{jiang2024mmsearch,
title={{MMSearch}: Benchmarking the potential of large models as multi-modal search engines},
author={Jiang, Dongzhi and Zhang, Renrui and Guo, Ziyu and Wu, Yanmin and Lei, Jiayi and Qiu, Pengshuo and Lu, Pan and Chen, Zehui and Fu, Chaoyou and Song, Guanglu and others},
journal={arXiv},
eprint={2409.12959},
archivePrefix={arXiv},
primaryClass={cs.IR},
year={2024}
}

@inproceedings{tang2025adaptive,
title={Adaptive keyframe sampling for long video understanding},
author={Tang, Xi and Qiu, Jihao and Xie, Lingxi and Tian, Yunjie and Jiao, Jianbin and Ye, Qixiang},
booktitle={Proceedings of the IEEE/CVF Conference on Computer Vision and Pattern Recognition},
pages={29118--29128},
year={2025}
}

@article{park2024too,
title={Too many frames, not all useful: Efficient strategies for long-form video {QA}},
author={Park, Jongwoo and Ranasinghe, Kanchana and Kahatapitiya, Kumara and Ryu, Wonjeong and Kim, Donghyun and Ryoo, Michael S.},
journal={arXiv},
eprint={2406.09396},
archivePrefix={arXiv},
primaryClass={cs.CV},
year={2024}
}

@inproceedings{liu2025bolt,
title={{BOLT}: Boost large vision-language model without training for long-form video understanding},
author={Liu, Shuming and Zhao, Chen and Xu, Tianqi and Ghanem, Bernard},
booktitle={Proceedings of the IEEE/CVF Conference on Computer Vision and Pattern Recognition},
pages={3318--3327},
year={2025}
}

@article{zhang2025q,
title={{Q}-Frame: Query-aware frame selection and multi-resolution adaptation for video-{LLM}s},
author={Zhang, Shaojie and Yang, Jiahui and Yin, Jianqin and Luo, Zhenbo and Luan, Jian},
journal={arXiv},
eprint={2506.22139},
archivePrefix={arXiv},
primaryClass={cs.CV},
year={2025}
}

@inproceedings{yu2024frame,
title={Frame-{Voyager}: Learning to query frames for video large language models},
author={Yu, Sicheng and Jin, Chengkai and Wang, Huanyu and Chen, Zhenghao and Jin, Sheng and Zuo, Zhongrong and Xu, Xiaolei and Sun, Zhenbang and Zhang, Bingni and Wu, Jiawei and others},
booktitle={International Conference on Learning Representations},
year={2025},
address={Singapore},
month={May}
}

@inproceedings{hu2025m,
title={{M}-{LLM}-based video frame selection for efficient video understanding},
author={Hu, Kai and Gao, Feng and Nie, Xiaohan and Zhou, Peng and Tran, Son and Neiman, Tal and Wang, Lingyun and Shah, Mubarak and Hamid, Raffay and Yin, Bing and others},
booktitle={Proceedings of the IEEE/CVF Conference on Computer Vision and Pattern Recognition},
pages={13702--13712},
year={2025}
}

@inproceedings{buch2025flexible,
title={Flexible frame selection for efficient video reasoning},
author={Buch, Shyamal and Nagrani, Arsha and Arnab, Anurag and Schmid, Cordelia},
booktitle={Proceedings of the IEEE/CVF Conference on Computer Vision and Pattern Recognition},
pages={29071--29082},
year={2025}
}

@inproceedings{yang2025pvc,
title={{PVC}: Progressive visual token compression for unified image and video processing in large vision-language models},
author={Yang, Chenyu and Dong, Xuan and Zhu, Xizhou and Su, Weijie and Wang, Jiahao and Tian, Hao and Chen, Zhe and Wang, Wenhai and Lu, Lewei and Dai, Jifeng},
booktitle={Proceedings of the IEEE/CVF Conference on Computer Vision and Pattern Recognition},
pages={24939--24949},
year={2025}
}

@article{shen2024longvu,
title={{LongVU}: Spatiotemporal adaptive compression for long video-language understanding},
author={Shen, Xiaoqian and Xiong, Yunyang and Zhao, Changsheng and Wu, Lemeng and Chen, Jun and Zhu, Chenchen and Liu, Zechun and Xiao, Fanyi and Varadarajan, Balakrishnan and Bordes, Florian and others},
journal={arXiv},
eprint={2410.17434},
archivePrefix={arXiv},
primaryClass={cs.CV},
year={2024}
}

@inproceedings{song2024moviechat,
title={MovieChat: From dense token to sparse memory for long video understanding},
author={Song, Enxin and Chai, Wenhao and Wang, Guanhong and Zhang, Yucheng and Zhou, Haoyang and Wu, Feiyang and Chi, Haozhe and Guo, Xun and Ye, Tian and Zhang, Yanting and others},
booktitle={Proceedings of the IEEE/CVF Conference on Computer Vision and Pattern Recognition},
pages={18221--18232},
year={2024}
}

@article{jin2025searchr1,
  title={Search-r1: Training llms to reason and leverage search engines with reinforcement learning},
  author={Jin, Bowen and Zeng, Hansi and Yue, Zhenrui and Yoon, Jinsung and Arik, Sercan and Wang, Dong and Zamani, Hamed and Han, Jiawei},
  journal={arXiv preprint arXiv:2503.09516},
  year={2025}
}

@article{liu2025visualarft,
title={Visual agentic reinforcement fine-tuning},
author={Liu, Ziyu and Zang, Yuhang and Zou, Yushan and Liang, Zijian and Dong, Xiaoyi and Cao, Yuhang and Duan, Haodong and Lin, Dahua and Wang, Jiaqi},
journal={arXiv},
eprint={2505.14246},
archivePrefix={arXiv},
primaryClass={cs.CV},
year={2025}
}

@article{pang2025mrvideo,
title={{Mr. Video}: {MapReduce} is the principle for long video understanding},
author={Pang, Ziqi and Wang, Yu-Xiong},
journal={arXiv},
eprint={2504.16082},
archivePrefix={arXiv},
primaryClass={cs.CV},
year={2025}
}

@article{zheng2025deepeyes,
title={DeepEyes: Incentivizing thinking with images via reinforcement learning},
author={Zheng, Ziwei and Yang, Michael and Hong, Jack and Zhao, Chenxiao and Xu, Guohai and Yang, Le and Shen, Chao and Yu, Xing},
journal={arXiv},
eprint={2505.14362},
archivePrefix={arXiv},
primaryClass={cs.CV},
year={2025}
}

@article{wu2025vtoolr1,
title={{VTool-R1}: {VLMs} learn to think with images via reinforcement learning on multimodal tool use},
author={Wu, Mingyuan and Yang, Jingcheng and Jiang, Jize and Li, Meitang and Yan, Kaizhuo and Yu, Hanchao and Zhang, Minjia and Zhai, Chengxiang and Nahrstedt, Klara},
journal={arXiv},
eprint={2505.19255},
archivePrefix={arXiv},
primaryClass={cs.CL},
year={2025}
}

@article{wang2025vrag-rl,
title={{VRAG-RL}: Empower vision-perception-based {RAG} for visually rich information understanding via iterative reasoning with reinforcement learning},
author={Wang, Qiuchen and Ding, Ruixue and Zeng, Yu and Chen, Zehui and Chen, Lin and Wang, Shihang and Xie, Pengjun and Huang, Fei and Zhao, Feng},
journal={arXiv},
eprint={2505.22019},
archivePrefix={arXiv},
primaryClass={cs.CV},
year={2025}
}

@article{wu2025mmsearchr1,
title={{MMSearch-R1}: Incentivizing {LMMs} to search},
author={Wu, Jinming and Deng, Zihao and Li, Wei and Liu, Yiding and You, Bo and Li, Bo and Ma, Zejun and Liu, Ziwei},
journal={arXiv},
eprint={2506.20670},
archivePrefix={arXiv},
primaryClass={cs.CL},
year={2025}
}

@article{xiao2025m2io-r1,
title={{M2IO-R1}: An efficient {RL}-enhanced reasoning framework for multimodal retrieval augmented multimodal generation},
author={Xiao, Zhiyou and Yu, Qinhan and Li, Binghui and Chen, Geng and Chen, Chong and Zhang, Wentao},
journal={arXiv},
eprint={2508.06328},
archivePrefix={arXiv},
primaryClass={cs.MM},
year={2025}
}

@article{cao2025ground,
  title = {{Ground}-{R1}: Incentivizing grounded visual reasoning via reinforcement learning},
  author = {Cao, Meng and Zhao, Haoze and Zhang, Can and Chang, Xiaojun and Reid, Ian and Liang, Xiaodan},
  journal = {arXiv},
  year = {2025},
  eprint = {2505.20272},
  primaryClass = {cs.CV}
}

@article{zhu2025active,
  title = {{Active}-{O3}: Empowering multimodal large language models with active perception via {GRPO}},
  author = {Zhu, Muzhi and Zhong, Hao and Zhao, Canyu and Du, Zongze and Huang, Zheng and Liu, Mingyu and Chen, Hao and Zou, Cheng and Chen, Jingdong and Yang, Ming and others},
  journal = {arXiv},
  year = {2025},
  eprint = {2505.21457},
  primaryClass = {cs.AI}
}

@article{zhang2025chain,
  title={{Chain-of-Focus}: Adaptive Visual Search and Zooming for Multimodal Reasoning via RL},
  author={Zhang, Xintong and Gao, Zhi and Zhang, Bofei and Li, Pengxiang and Zhang, Xiaowen and Liu, Yang and Yuan, Tao and Wu, Yuwei and Jia, Yunde and Zhu, Song-Chun and others},
  journal={arXiv preprint arXiv:2505.15436},
  year={2025}
}

@article{su2025pixel,
  title = {{Pixel Reasoner}: Incentivizing pixel-space reasoning with curiosity-driven reinforcement learning},
  author = {Su, Alex and Wang, Haozhe and Ren, Weiming and Lin, Fangzhen and Chen, Wenhu},
  journal = {arXiv},
  year = {2025},
  eprint = {2505.15966},
  primaryClass = {cs.AI}
}

@article{jiang2025vlm,
  title = {{VLM}-{R$^{3}$}: Region recognition, reasoning, and refinement for enhanced multimodal chain-of-thought},
  author = {Jiang, Chaoya and Heng, Yongrui and Ye, Wei and Yang, Han and Xu, Haiyang and Yan, Ming and Zhang, Ji and Huang, Fei and Zhang, Shikun},
  journal = {arXiv},
  year = {2025},
  eprint = {2505.16192},
  primaryClass = {cs.CV}
}

@article{zhang2025thyme,
  title = {{Thyme}: Think beyond images},
  author = {Zhang, Yi-Fan and Lu, Xingyu and Yin, Shukang and Fu, Chaoyou and Chen, Wei and Hu, Xiao and Wen, Bin and Jiang, Kaiyu and Liu, Changyi and Zhang, Tianke and others},
  journal = {arXiv},
  year = {2025},
  eprint = {2508.11630},
  primaryClass = {cs.CV}
}

@article{su2025openthinkimg,
  title = {{OpenThinkIMG}: Learning to think with images via visual tool reinforcement learning},
  author = {Su, Zhaochen and Li, Linjie and Song, Mingyang and Hao, Yunzhuo and Yang, Zhengyuan and Zhang, Jun and Chen, Guanjie and Gu, Jiawei and Li, Juntao and Qu, Xiaoye and others},
  journal = {arXiv},
  year = {2025},
  eprint = {2505.08617},
  primaryClass = {cs.CV}
}

@article{wu2025reinforcing,
  title = {Reinforcing spatial reasoning in vision-language models with interwoven thinking and visual drawing},
  author = {Wu, Junfei and Guan, Jian and Feng, Kaituo and Liu, Qiang and Wu, Shu and Wang, Liang and Wu, Wei and Tan, Tieniu},
  journal = {arXiv},
  year = {2025},
  eprint = {2506.09965},
  primaryClass = {cs.CV}
}

@article{xu2025visual,
  title = {Visual {Planning}: Let's think only with images},
  author = {Xu, Yi and Li, Chengzu and Zhou, Han and Wan, Xingchen and Zhang, Caiqi and Korhonen, Anna and Vuli{\'c}, Ivan},
  journal = {arXiv},
  year = {2025},
  eprint = {2505.11409},
  primaryClass = {cs.CL}
}

@article{ren2025videorag,
  title = {{VideoRAG}: Retrieval-augmented generation with extreme long-context videos},
  author = {Ren, Xubin and Xu, Lingrui and Xia, Long and Wang, Shuaiqiang and Yin, Dawei and Huang, Chao},
  journal = {arXiv},
  year = {2025},
  eprint = {2502.01549},
  primaryClass = {cs.CL}
}

@inproceedings{wang2025video,
  title = {{Video-RTS}: Rethinking reinforcement learning and test-time scaling for efficient and enhanced video reasoning},
  author = {Wang, Ziyang and Yoon, Jaehong and Yu, Shoubin and Islam, Md Mohaiminul and Bertasius, Gedas and Bansal, Mohit},
  booktitle = {Proceedings of the Conference on Empirical Methods in Natural Language Processing},
  pages = {28114--28128},
  year = {2025}
}

@article{guo2025deepseek,
  title = {{DeepSeek-R1}: Incentivizing reasoning capability in {LLMs} via reinforcement learning},
  author = {Guo, Daya and Yang, Dejian and Zhang, Haowei and Song, Junxiao and Zhang, Ruoyu and Xu, Runxin and Zhu, Qihao and Ma, Shirong and Wang, Peiyi and Bi, Xiao and others},
  journal = {arXiv},
  year = {2025},
  eprint = {2501.12948},
  primaryClass = {cs.LG},
  note = {Also known as {DeepSeek-R1}}
}

@article{zhang2025deep,
  title = {{Deep Video Discovery}: Agentic search with tool use for long-form video understanding},
  author = {Zhang, Xiaoyi and Jia, Zhaoyang and Guo, Zongyu and Li, Jiahao and Li, Bin and Li, Houqiang and Lu, Yan},
  journal = {arXiv},
  year = {2025},
  eprint = {2505.18079},
  primaryClass = {cs.CV}
}

@inproceedings{wang2025adaretake,
  title = {{AdaReTake}: Adaptive redundancy reduction to perceive longer for video-language understanding},
  author = {Wang, Xiao and Si, Qingyi and Zhu, Shiyu and Wu, Jianlong and Cao, Li and Nie, Liqiang},
  booktitle = {Findings of the Association for Computational Linguistics: {ACL} {2025}},
  pages = {5417--5432},
  year = {2025}
}

@article{zhang2025flexselect,
  title = {{FlexSelect}: Flexible token selection for efficient long video understanding},
  author = {Zhang, Yunzhu and Lu, Yu and Wang, Tianyi and Rao, Fengyun and Yang, Yi and Zhu, Linchao},
  journal = {arXiv},
  year = {2025},
  eprint = {2506.00993},
  primaryClass = {cs.CV}
}

@inproceedings{fu2025videomme,
  title = {{Video-MME}: the first-ever comprehensive evaluation benchmark of multimodal {LLMs} in video analysis},
  author = {Fu, Chaoyou and Dai, Yuhan and Luo, Yongdong and Li, Lei and Ren, Shuhuai and Zhang, Renrui and Wang, Zihan and Zhou, Chenyu and Shen, Yunhang and Zhang, Mengdan and others},
  booktitle = {Proceedings of the IEEE/CVF Conference on Computer Vision and Pattern Recognition},
  pages = {24108--24118},
  year = {2025}
}

@inproceedings{wang2025lvbench,
  title = {{LVBench}: an extreme long video understanding benchmark},
  author = {Wang, Weihan and He, Zehai and Hong, Wenyi and Cheng, Yean and Zhang, Xiaohan and Qi, Ji and Ding, Ming and Gu, Xiaotao and Huang, Shiyu and Xu, Bin and others},
  booktitle = {Proceedings of the IEEE/CVF International Conference on Computer Vision},
  pages = {22958--22967},
  year = {2025}
}

@article{wu2024longvideobench,
  title = {{LongVideoBench}: a benchmark for long-context interleaved video-language understanding},
  author = {Wu, Haoning and Li, Dongxu and Chen, Bei and Li, Junnan},
  journal = {Advances in Neural Information Processing Systems},
  volume = {37},
  pages = {28828--28857},
  year = {2024}
}

@article{parr2017uncertainty,
  title = {Uncertainty, epistemics and active inference},
  author = {Parr, Thomas and Friston, Karl J},
  journal = {Journal of the Royal Society Interface},
  volume = {14},
  number = {136},
  pages = {20170376},
  year = {2017},
  publisher = {The Royal Society}
}

@article{alexander2015epistemic,
  title = {Epistemic landscapes, optimal search, and the division of cognitive labor},
  author = {Alexander, Jason McKenzie and Himmelreich, Johannes and Thompson, Christopher},
  journal = {Philosophy of Science},
  volume = {82},
  number = {3},
  pages = {424--453},
  year = {2015},
  publisher = {Cambridge University Press}
}

@inproceedings{chandrasekharan2004reactive,
  title = {Reactive agents learn to add epistemic structures to the world},
  author = {Chandrasekharan, Sanjay and Stewart, Terry},
  booktitle = {Proceedings of the Annual Meeting of the Cognitive Science Society},
  volume = {26},
  number = {26},
  year = {2004}
}

@article{hurst2024gpt,
  title={Gpt-4o system card},
  author={Hurst, Aaron and Lerer, Adam and Goucher, Adam P and Perelman, Adam and Ramesh, Aditya and Clark, Aidan and Ostrow, AJ and Welihinda, Akila and Hayes, Alan and Radford, Alec and others},
  journal={arXiv preprint arXiv:2410.21276},
  year={2024}
}

@article{bai2025qwen2,
  title = {{Qwen2.5-VL} technical report},
  author = {Bai, Shuai and Chen, Keqin and Liu, Xuejing and Wang, Jialin and Ge, Wenbin and Song, Sibo and Dang, Kai and Wang, Peng and Wang, Shijie and Tang, Jun and others},
  journal = {arXiv},
  eprint = {2502.13923},
  archivePrefix = {arXiv},
  year = {2025}
}

@article{liu2023visual,
  title = {Visual instruction tuning},
  author = {Liu, Haotian and Li, Chunyuan and Wu, Qingyang and Lee, Yong Jae},
  journal = {Advances in Neural Information Processing Systems},
  volume = {36},
  pages = {34892--34916},
  year = {2023}
}

@misc{OpenAI2025o3o4mini,
  author       = {{OpenAI}},
  title        = {Introducing {OpenAI} o3 and o4-mini},
  howpublished = {\url{https://openai.com/index/introducing-o3-and-o4-mini/}},
  year         = {2025},
}

@article{tian2025ego-r1,
  title = {{Ego-R1}: {Chain-of-Tool-Thought} for ultra-long egocentric video reasoning},
  author = {Tian, Shulin and Wang, Ruiqi and Guo, Hongming and Wu, Penghao and Dong, Yuhao and Wang, Xiuying and Yang, Jingkang and Zhang, Hao and Zhu, Hongyuan and Liu, Ziwei},
  journal = {arXiv},
  eprint = {2506.13654},
  archivePrefix = {arXiv},
  year = {2025}
}

@article{long2025seeing,
  title = {Seeing, listening, remembering, and reasoning: A multimodal agent with long-term memory},
  author = {Long, Lin and He, Yichen and Ye, Wentao and Pan, Yiyuan and Lin, Yuan and Li, Hang and Zhao, Junbo and Li, Wei},
  journal = {arXiv},
  eprint = {2508.09736},
  archivePrefix = {arXiv},
  year = {2025}
}

@inproceedings{wang2024videoagent,
  title = {{VideoAgent}: Long-form video understanding with large language model as agent},
  author = {Wang, Xiaohan and Zhang, Yuhui and Zohar, Orr and Yeung-Levy, Serena},
  booktitle = {Proceedings of the European Conference on Computer Vision},
  pages = {58--76},
  year = {2024},
  publisher = {Springer}
}

@article{meng2025cyberv,
  title = {{CyberV}: Cybernetics for test-time scaling in video understanding},
  author = {Meng, Jiahao and Sun, Shuyang and Tan, Yue and Qi, Lu and Tong, Yunhai and Li, Xiangtai and Wen, Longyin},
  journal = {arXiv},
  eprint = {2506.07971},
  archivePrefix = {arXiv},
  year = {2025}
}

@article{zhou2025reagent,
  title = {{ReAgent-V}: A reward-driven multi-agent framework for video understanding},
  author = {Zhou, Yiyang and He, Yangfan and Su, Yaofeng and Han, Siwei and Jang, Joel and Bertasius, Gedas and Bansal, Mohit and Yao, Huaxiu},
  journal = {arXiv},
  eprint = {2506.01300},
  archivePrefix = {arXiv},
  year = {2025}
}

@article{feng2025video,
  title = {{Video-R1}: Reinforcing video reasoning in {MLLMs}},
  author = {Feng, Kaituo and Gong, Kaixiong and Li, Bohao and Guo, Zonghao and Wang, Yibing and Peng, Tianshuo and Wu, Junfei and Zhang, Xiaoying and Wang, Benyou and Yue, Xiangyu},
  journal = {arXiv},
  eprint = {2503.21776},
  archivePrefix = {arXiv},
  year = {2025}
}

@inproceedings{suris2023vipergpt,
  title = { {ViperGPT}: Visual inference via {Python} execution for reasoning},
  author = {Sur{\'\i}s, D{\'\i}dac and Menon, Sachit and Vondrick, Carl},
  booktitle = {Proceedings of the {IEEE}/{CVF} International Conference on Computer Vision},
  pages = {11888--11898},
  year = {2023}
}

@inproceedings{zhao2025mmvu,
  title = {{MMVU}: Measuring expert-level multi-discipline video understanding},
  author = {Zhao, Yilun and Zhang, Haowei and Xie, Lujing and Hu, Tongyan and Gan, Guo and Long, Yitao and Hu, Zhiyuan and Chen, Weiyuan and Li, Chuhan and Xu, Zhijian and others},
  booktitle = {Proceedings of the {IEEE}/{CVF} Conference on Computer Vision and Pattern Recognition},
  pages = {8475--8489},
  year = {2025}
}

@article{hu2025video,
  title   = {Video-{MMMU}: Evaluating Knowledge Acquisition from Multi-Discipline Professional Videos},
  author  = {Hu, Kairui and Wu, Penghao and Pu, Fanyi and Xiao, Wang and Zhang, Yuanhan and Yue, Xiang and Li, Bo and Liu, Ziwei},
  journal = {arXiv preprint arXiv:2501.13826},
  year    = {2025}
}

@inproceedings{mangalam2023egoschema,
  title     = {{EgoSchema}: A Diagnostic Benchmark for Very Long-Form Video Language Understanding},
  author    = {Mangalam, Karttikeya and Akshulakov, Raiymbek and Malik, Jitendra},
  booktitle = {Advances in Neural Information Processing Systems},
  volume    = {36},
  pages     = {46212--46244},
  year      = {2023}
}

@misc{openai2025gpt5,
  author  = {{OpenAI}},
  title   = {{GPT-5} Is Here},
  year    = {2025},
  url     = {https://openai.com/gpt-5/},
  urldate = {2025-08-07}
}

@misc{openai2025gpt4.1,
  author  = {{OpenAI}},
  title   = {Introducing {GPT-4.1} in the {API}},
  year    = {2025},
  url     = {https://openai.com/index/gpt-4-1/},
  urldate = {2025-04-14}
}

@article{GoogleDeepMind2025Gemini25,
  title   = {Gemini 2.5: Pushing the Frontier with Advanced Reasoning, Multimodality, Long Context, and Next Generation Agentic Capabilities},
  author  = {Comanici, Gheorghe and Bieber, Eric and Schaekermann, Mike and Pasupat, Ice and Sachdeva, Noveen and Dhillon, Inderjit and Blistein, Marcel and Ram, Ori and Zhang, Dan and Rosen, Evan and others},
  journal = {arXiv preprint arXiv:2507.06261},
  year    = {2025}
}

@inproceedings{yang2025vca,
  title     = {{VCA}: Video Curious Agent for Long Video Understanding},
  author    = {Yang, Zeyuan and Chen, Delin and Yu, Xueyang and Shen, Maohao and Gan, Chuang},
  booktitle = {Proceedings of the {IEEE}/{CVF} International Conference on Computer Vision},
  pages     = {20168--20179},
  year      = {2025}
}

@article{team2024gemini1.5,
  title   = {Gemini 1.5: Unlocking Multimodal Understanding Across Millions of Tokens of Context},
  author  = {Team, Gemini and Georgiev, Petko and Lei, Ving Ian and Burnell, Ryan and Bai, Libin and Gulati, Anmol and Tanzer, Garrett and Vincent, Damien and Pan, Zhufeng and Wang, Shibo and others},
  journal = {arXiv preprint arXiv:2403.05530},
  year    = {2024}
}

@misc{GoogleDeepMind2024Gemini2,
  author  = {{Google DeepMind}},
  title   = {Introducing Gemini 2.0: Our New {AI} Model for the Agentic Era},
  year    = {2024},
  url     = {https://blog.google/technology/google-deepmind/google-gemini-ai-update-december-2024/},
  urldate = {2024-12-XX}
}

@misc{
zhang2024video,
title={Video Instruction Tuning with Synthetic Data},
author={Yuanhan Zhang and Jinming Wu and Wei Li and Bo Li and Zejun MA and Ziwei Liu and Chunyuan Li},
year={2024},
url={https://openreview.net/forum?id=8Livf4oZxz}
}

@article{chen2024internvl2.5,
  title={Expanding performance boundaries of open-source multimodal models with model, data, and test-time scaling},
  author={Chen, Zhe and Wang, Weiyun and Cao, Yue and Liu, Yangzhou and Gao, Zhangwei and Cui, Erfei and Zhu, Jinguo and Ye, Shenglong and Tian, Hao and Liu, Zhaoyang and others},
  journal={arXiv preprint arXiv:2412.05271},
  year={2024}
}

@article{chen2024internvl2,
  title={How far are we to gpt-4v? closing the gap to commercial multimodal models with open-source suites},
  author={Chen, Zhe and Wang, Weiyun and Tian, Hao and Ye, Shenglong and Gao, Zhangwei and Cui, Erfei and Tong, Wenwen and Hu, Kongzhi and Luo, Jiapeng and Ma, Zheng and others},
  journal={Science China Information Sciences},
  volume={67},
  number={12},
  pages={220101},
  year={2024},
  publisher={Springer}
}

@article{li2024aria,
  title={Aria: An open multimodal native mixture-of-experts model},
  author={Li, Dongxu and Liu, Yudong and Wu, Haoning and Wang, Yue and Shen, Zhiqi and Qu, Bowen and Niu, Xinyao and Zhou, Fan and Huang, Chengen and Li, Yanpeng and others},
  journal={arXiv preprint arXiv:2410.05993},
  year={2024}
}

@article{yang2025qwen3,
  title={Qwen3 technical report},
  author={Yang, An and Li, Anfeng and Yang, Baosong and Zhang, Beichen and Hui, Binyuan and Zheng, Bo and Yu, Bowen and Gao, Chang and Huang, Chengen and Lv, Chenxu and others},
  journal={arXiv preprint arXiv:2505.09388},
  year={2025}
}

@misc{zai2025glm47,
  author       = {{Z.ai}},
  title        = {{GLM-4.7: Advancing the Coding Capability}},
  year         = {2025},
  month        = dec,
  howpublished = {\url{https://z.ai/blog/glm-4.7}},
  note         = {Official blog post, accessed March 18, 2026}
}

@misc{deepseek2025deepseekv3,
  author       = {{DeepSeek-AI}},
  title        = {{DeepSeek-V3}},
  year         = {2025},
  howpublished = {\url{https://github.com/deepseek-ai/DeepSeek-V3}},
  note         = {Official GitHub repository, accessed March 18, 2026}
}
}

\appendix
\maketitlesupplementary
\section{Method Details}
\label{app:method-details}

This appendix provides detailed designs of the LensWalk agent, including the overall reason--plan--observe algorithm used in our framework, the concrete prompts for each module, and tool schemas.

\subsection{Algotithm Demonstration}
\label{sec:agent-alg-memory}
 
We present abstract pseudocode for the LensWalk reason--plan--observe loop in Algorithm \ref{alg:lenswalk}. Using the notation of Section~\ref{sec:method}, it shows how each Reasoner plan $a_t = (o_t, q_t, \mathcal{I}_t, \rho_{o_t})$ over tools $\mathcal{O} = \{\textit{Scan Search}, \textit{Segment Focus}, \textit{Stitched Verify}, \textit{Finish}\}$ is translated into one or more VLM-visible context groups, how the Observer $M_o$ is queried on those groups, and how the resulting time-anchored evidence and subject table state $Sub_t$ are threaded through the multi-turn loop.
\begin{algorithm*}[!ht]
    \caption{LensWalk Reason--Plan-Observe loop with Subject Table Memory update.}
    \label{alg:lenswalk}
    \SetKwInOut{Input}{Input}
    \SetKwInOut{Output}{Output}
    \Input{Video $V$, query $Q$, metadata $V_{\text{info}}$, tool set $\mathcal{O} = \{\textit{Scan Search}, \textit{Segment Focus}, \textit{Stitched Verify}, \textit{Finish}\}$, max steps $T_{\max}$}
    \Output{Answer to $Q$}
    Initialize history $L_0 \leftarrow \{Q, V_{\text{info}}\}$\;
    Initialize subject table $Sub_0 \leftarrow \emptyset$\;
    \For{$t = 1, 2, \dots, T_{\max}$}{
        $a_t = (o_t, q_t, \mathcal{I}_t, \rho_{o_t}) \leftarrow M_r(Q, V_{\text{info}}, L_{t-1}, Sub_{t-1})$\;
        \If{$o_t = \textit{Finish}$}{%
            \Return{$q_t$ as final answer}\;
        }
        $G_t \leftarrow \mathrm{BuildContextGroups}(o_t, \mathcal{I}_t, \rho_{o_t})$\tcp*[r]{Maps tool-level sampling arguments into VLM-visible context groups}
        \tcp{$G_t = \{(I_t^{(k)}, s_t^{(k)})\}_{k=1}^{K_t}$, where $I_t^{(k)}$ is the $k$-th time interval and $s_t^{(k)}$ collects its sampling settings (e.g., fps, frame budget)}
        \tcp{$K_t = 1$ for \textit{Segment Focus} and \textit{Stitched Verify}; $K_t > 1$ (parallel slices) for \textit{Scan Search}}
        \For{each $(I_t^{(k)}, s_t^{(k)}) \in G_t$}{
            $F_t^{(k)} \leftarrow \mathrm{SampleFrames}(V, I_t^{(k)}, s_t^{(k)})$\tcp*[r]{Extracts frames in $I_t^{(k)}$ from $V$ according to sampling config $s_t^{(k)}$}
            $y_t^{(k)} \leftarrow M_o(q_t, F_t^{(k)}, I_t^{(k)})$\;
        }
        $e_t \leftarrow \mathrm{Aggregate}(\{y_t^{(k)}\}_{k=1}^{K_t})$\tcp*[r]{Simple join: concatenate per-group replies in temporal order into a single observation string}
        $L_t \leftarrow L_{t-1} \cup \{(a_t, e_t)\}$\;
        \If{subject memory enabled}{
            $Sub_t \leftarrow \mathrm{MemoryUpdate}(Sub_{t-1}, a_t, e_t)$\;
        }
        \Else{
            $Sub_t \leftarrow Sub_{t-1}$\;
        }
    }
    \Return{unknown}\;
\end{algorithm*}
\subsection{Prompt Design}
\label{app:prompt-design}

We list the main prompts used by the planning \textit{Reasoner} and the tool-specific \textit{Observer} models. All prompts are shown verbatim as they are instantiated in our implementation.

%\subsubsection{Reasoner Prompt}
\paragraph{Reasoner Prompt.}
\label{app:reasoner-system-prompt}
The dedicated system prompt driving the multi-turn planning Reasoner is given in \cref{fig:reasoner-system-prompt}. It defines the THINK--PLAN--OBSERVE loop, the tool-usage guidelines, and the requirement to terminate via the \texttt{finish} tool. We pair it with the user prompt used in our experiments, as in \cref{fig:reasoner-user-prompt}.

%\subsubsection{Tool-Call Decision Output Schema}
\paragraph{Tool-Call Decision Output Schema.}
\label{app:tool-call-schema}

At each reasoning turn, the Reasoner emits a tool call in the format natively defined by the OpenAI official SDK\cite{hurst2024gpt}. Concretely, its raw response contains a \texttt{tool\_calls} list where each element has the standard structure
\texttt{
\{ "type": "function", "function": \{"name": `name of the tool', "arguments": `tool arguments'\} \}
}.
This format would be parsed the \texttt{function.name} and JSON-stringify \texttt{function.arguments} fields into a Python dictionary before dispatching to execute this tool.

%\subsubsection{Observer Prompts (per Tool)}
\paragraph{Observer Prompts (per Tool).}
\label{app:observer-prompts}

Each video-observation tool is paired with its own Observer prompt. We show one figure of each tool, such as ``Scan Search'', ``Segment Focus'', and ``Stitch Verify'', in Figure \ref{fig:scan-observer-prompt}, \ref{fig:single-observer-prompt}, \ref{fig:stitched-observer-prompt}, respectively.

%\subsubsection{Subject Memory Update Prompts}
\paragraph{Subject Memory Update Prompts.}\label{app:subject-memory-update-prompts}

When subject-level memory is enabled, the agent uses a dedicated memory-update prompt to reconcile subject information from per-step observations with a global subject registry. The concrete memory-update prompt is described in Figure~\ref{fig:memory-update-prompt}.

\subsection{Tool Schema, Format, and Signatures}
\label{app:tool-schema}

We next describe the tool schema exposed to the Reasoner. Tool calls of Reasoner must follow the given JSON schema. It is provided with a tool \texttt{name}, a short natural-language \texttt{description}, and an \texttt{arguments} object whose fields include names, types, optionality, and brief semantic descriptions. The full instantiation for the LensWalk tools is shown in \cref{fig:tool-signatures}.
\section{Evaluation Details}
\label{sec:Evaluation Details}
\subsection{Implementation Details}
\paragraph{Models.}
Table~\ref{tab:models_used} lists the models used as Reasoner/Observer in our experiments in Section \ref{sec:exp}. ``Releases'' specifies the time of the first public API or open-weight announcement, and ''Version'' column corresponds to the exact alias used in our experiments.

\paragraph{LensWalk Agent settings.}
We restrict the dialogue to at most 20 tool-calling turns. For the budgets of three observation tools, \textit{Scan Search} samples 30 frames per slice with a default sampling rate of 0.25\,fps; \textit{Segment Focus} defaults to 1\,fps with at most 32 frames for the requested interval; \textit{Stitch Verify} stitches plan-specified segments, defaulting to 0.5\,fps globally and 1\,fps per segment with an overall cap of 128 frames. Requested sampling in each tool call is first converted from the provided (interval, fps) pairs to a frame budget and then proportionally rescaled so the total does not exceed each tool’s \texttt{max\_total\_frame}. 

\paragraph{Deployment and Service.}
Closed-source models (o3, GPT-4.1/4.1-mini/5) and Qwen3-235B-A22B are queried through their official APIs. All other open-weight models, including Qwen2.5-VL-72B/7B, are served with vLLM Engine on 4{\small$\times$}NVIDIA H100 80\,GB GPUs.
\subsection{Benchmark Datasets}
We assess LensWalk on six long-form or reasoning-focused benchmarks. \textbf{LVBench}\cite{wang2025lvbench} provides 1{,}549 multiple-choice questions drawn from 103 curated hour-long movies (mean duration 67\,min); transcripts are unavailable, so the agent relies on frames only. \textbf{LongVideoBench-long}\cite{wu2024longvideobench} contributes 564 MCQs sourced from 188 validation videos in the 900--3{,}600\,s bucket; this split is the only one where we feed automatic transcripts to the Reasoner, because the benchmark ships ASR JSON files. \textbf{Video-MME long}\cite{fu2025videomme} is the long split of Video-MME covers 900 MCQs over 300 videos lasting 30--60\,min (mean 41\,min) across knowledge, skill, and perception domains; we keep the official multiple-choice format and do not use subtitles for evaluation. For \textbf{MMVU}\cite{zhao2025mmvu}, 625 multiple-choice questions from the validation set spanning 40+ academic subjects are used. \textbf{Video-MMMU}\cite{hu2025video} aggregates 900 MCQs evenly split across the Perception, Comprehension, and Adaptation subsets, each averaging 8.4-minute clips ranging from short 48\,s labs to 30+ minute lectures. \textbf{EgoSchema}\cite{mangalam2023egoschema} supplies the official validation set of 500 egocentric three-minute clips with 500 multiple-choice questions.

\subsection{Baseline Reproduction}
Most baseline results are obtained from their official papers or public leaderboards (e.g., Deep Video Discovery\cite{zhang2025deep}, Mr. Video\cite{pang2025mrvideo}, VideoAgent\cite{wang2024videoagent}, Ego-R1\cite{tian2025ego-r1}). We only re-run GPT-4.1, GPT-5 on the considered benchmarks, and Qwen2.5-VL-72B on LongVideoBench-long and EgoSchema-val. For GPT-5, we fix \textit{reasoning\_effort} to \texttt{minimal} both inside our agent and in the baseline setting. Frame sampling budgets follow prior works: GPT-4.1/5 consume 256 frames, while local Qwen2.5-VL-72B is limited to 128 frames due to our limited computing resources, matching the max per-forward cap of the \textit{observer} tool.
\begin{figure*}[!ht]
    \centering
    \begin{tcolorbox}[enhanced,breakable,colback=gray!10,colframe=gray!75,arc=3mm,boxrule=0.5pt]
    \small
    \textbf{You are a helpful assistant designed to answer multi-step questions about video content by sequentially invoking tools for video observation.} Your sole purpose is to reason and act to uncover the correct answer. Operate in a rigorous \textbf{THINK $\rightarrow$ PLAN $\rightarrow$ OBSERVE} loop with at most \texttt{MAX\_CALL} tool calls.

    \textbf{THINK}: Based on the user's query and all previous observations, reason step by step about what is known, what is uncertain, and what to do next. Justify the next action by stating the goal, the best tool, chosen time ranges, and the sampling strategy (e.g., higher FPS for fast actions, lower for context).

    \textbf{ACT}: Call exactly one video observation tool with the decided parameters. Time ranges and sampling intensity should match the planned evidence needs, avoiding both information loss and redundancy.

    \textbf{OBSERVE}: Use the returned structured observation to update understanding and continue the next THINK cycle.

    Continue until you have sufficient evidence, then call \texttt{finish} with the final answer.

    \textbf{Core directives}: (1) Never guess; inspect the video when uncertain. (2) Plan within frame budget. (3) Iterate and refine. (4) Verify and synthesize after multiple related segments. (5) Ground arguments in evidence; do not invent timestamps.
    \end{tcolorbox}
    \caption{System prompt used for the LensWalk \textit{Reasoner}. }
    \label{fig:reasoner-system-prompt}
\end{figure*}

\begin{figure*}[!ht]
    \centering
    \begin{tcolorbox}[enhanced,breakable,colback=gray!10,colframe=gray!75,arc=3mm,boxrule=0.5pt]
    \small
    % \textbf{VCS\_REASON\_USER\_PROMPT} \\[0.2em]
    Based on your observations and tool outputs, provide a concise answer that directly addresses the question. Only provide the option's letter from the given choices as the final answer for multiple-choice questions.\\[0.4em]
    Total video length: \texttt{VIDEO\_LENGTH} seconds.\\
    Question: \texttt{QUESTION\_PLACEHOLDER}
    \end{tcolorbox}
    \caption{User prompt used for LensWalk \textit{Reasoner}.}
    \label{fig:reasoner-user-prompt}
\end{figure*}
\begin{figure*}[!ht]
    \centering
    \begin{tcolorbox}[enhanced,breakable,colback=gray!10,colframe=gray!75,arc=3mm,boxrule=0.5pt]
    \small
    You are a highly focused visual understanding assistant. Your task is to analyze a short sequence of sparsely sampled video frames from a single segment and assess its relevance to the user's query. You need to provide a concise summary of the visual content and explain why it is, or is not, related to the query.\\[0.3em]
    \textbf{IMPORTANT:} Your scope is strictly limited to the provided frames from this single segment. Your conclusion must be based only on what is visible here. Do not make any assumptions about the rest of the video or events outside these frames. Your response should be a brief, self-contained analysis of this segment's relevance.
    \end{tcolorbox}
    \caption{Prompt for the \textit{Observer} in \textit{Scan Search} tool.}
    \label{fig:scan-observer-prompt}
\end{figure*}

\begin{figure*}[!ht]
    \centering
    \begin{tcolorbox}[enhanced,breakable,colback=gray!10,colframe=gray!75,arc=3mm,boxrule=0.5pt]
    \small
    You are a meticulous and factual video analysis assistant. Your task is to analyze a set of sparsely sampled frames from a single, continuous time segment of a video. Provide a detailed and objective description that directly answers the user's query.\\[0.3em]
    \textbf{IMPORTANT:} Base your entire analysis strictly on the visual information present in the frames provided. The frames are sampled uniformly. Do not infer or imagine events happening beyond the frames. Your output should be a concise and structured text focused solely on the evidence at hand.
    \end{tcolorbox}
    \caption{Prompt for the \textit{Observer} in \textit{Segment Focus} tool.}
    \label{fig:single-observer-prompt}
\end{figure*}

\begin{figure*}[!ht]
    \centering
    \begin{tcolorbox}[enhanced,breakable,colback=gray!10,colframe=gray!75,arc=3mm,boxrule=0.5pt]
    \small
    You are an expert video analyst specializing in synthesizing information from multiple, distinct key moments. You will be shown a ``stitched'' collection of frames sampled from various, potentially non-contiguous time intervals of a video. Your goal is to provide a cohesive analysis that addresses the user's query, drawing connections or noting changes between the different moments shown.\\[0.3em]
    \textbf{IMPORTANT:} The frames represent a curated, sparse, and non-uniform selection. Do not assume what happens in the time gaps between the provided segments. Base your entire analysis strictly on the visual evidence presented, treating each time-stamped group as a distinct snapshot.
    \end{tcolorbox}
    \caption{Prompt for the \textit{Observer} in  \textit{Stitched Verify} tool (\texttt{stitched\_observer}).}
    \label{fig:stitched-observer-prompt}
\end{figure*}

\begin{figure*}[!ht]
    \centering
    \begin{tcolorbox}[enhanced,breakable,colback=gray!10,colframe=gray!75,arc=3mm,boxrule=0.5pt]
    \small
    \textbf{Memory role.} You are a specialized assistant responsible for maintaining a \texttt{subject\_registry} for a video analysis agent. You analyze the latest observation and its subjects, compare them against the existing registry, and produce a consolidated registry only (no narrative summary).

    \textbf{Input provided.}
    \begin{itemize}
        \item Current subject registry (master list so far).
        \item History: past tool calls and their results.
        \item Current turn: latest tool call, its raw output, and a \texttt{[new observed subject registry]} extracted from that call.
    \end{itemize}

    \textbf{Update rules.}
    \begin{enumerate}
        \item \textbf{Merge by synthesis, not stacking.} When updating a subject's description list, synthesize concise, high-value descriptions instead of appending redundant strings. Example: from ``man in red shirt'' and ``man in red shirt, sitting at desk'' produce ``man in red shirt'' and ``sitting at desk'' (or ``man in red shirt (was standing, now sitting at desk)''), not duplicated phrases.
        \item \textbf{Add new subjects} when they cannot be merged.
        \item \textbf{Prune to the 15 most recently observed.} After merge/add, keep at most 15 subjects prioritized by most recent appearance, giving current-turn subjects highest priority.
        \item \textbf{No subjects} $\Rightarrow$ output an empty object \{\}.
    \end{enumerate}

    \textbf{Output format (single JSON object).}
    \begin{lstlisting}[linewidth=\linewidth,breaklines=true,breakatwhitespace=true,keepspaces=true]
{
  "updated_subject_registry": {
    "A unique and consistent identifier for a subject": {
      "description": [
        "Key appearance/role/state strings (e.g., 'man in red shirt', 'sitting at desk', 'talking to S2')."
      ],
      "appeared_intervals": [
        "[start_sec, end_sec]"
      ]
    }
  }
}
    \end{lstlisting}
    \end{tcolorbox}
    \caption{Subject Table Memory--Update prompt and required JSON output.}
    \label{fig:memory-update-prompt}
\end{figure*}
\begin{table*}[!t]
    \centering
    \caption{\small Models used in our experiments.}
    \label{tab:models_used}
    \setlength{\tabcolsep}{5pt}
    \renewcommand{\arraystretch}{1.0}
    \begin{tabular}{cllr}
        \toprule
        \textbf{Organization} & \textbf{Model} & \textbf{Release} & \textbf{Version} \\
        \midrule
        \multirow{4}{*}{OpenAI}
            & GPT-4.1\cite{openai2025gpt4.1} & 2025-04 & \texttt{gpt-4.1-2025-04-14} \\
            & GPT-4.1-mini\cite{openai2025gpt4.1} & 2025-04 & \texttt{gpt-4.1-mini-2025-04-14} \\
            & o3\cite{OpenAI2025o3o4mini} & 2025-04 & \texttt{o3-2025-04-16} \\
            & GPT-5\cite{openai2025gpt5} & 2025-08 & \texttt{gpt-5-2025-08-07} \\ %
        \midrule
        \multirow{3}{*}{Alibaba}
            & Qwen3-235B-A22B\cite{yang2025qwen3} & 2025-04 & \texttt{qwen3-235b-a22b-instruct} \\
            & Qwen2.5-VL-72B\cite{bai2025qwen2} & 2025-02 & \texttt{qwen2.5-vl-72b-instruct} \\
            & Qwen2.5-VL-7B\cite{bai2025qwen2}  & 2025-02 & \texttt{qwen2.5-vl-7b-instruct} \\
        \midrule
            \multirow{1}{*}{Z.ai}
                & GLM-4.7\cite{zai2025glm47} & 2025-08 & \texttt{glm-4.7} \\
        \midrule
        \multirow{1}{*}{DeepSeek}
            & DeepSeek-V3\cite{deepseek2025deepseekv3} & 2025-03 & \texttt{DeepSeek-V3-0324} \\
                \bottomrule
    \end{tabular}
    \vspace{-6pt}
\end{table*}

\begin{figure*}[t]
    \centering
    \begin{tcolorbox}[enhanced,breakable,colback=gray!10,colframe=gray!75,
                      arc=3mm,boxrule=0.5pt]
\linespread{0.8}\selectfont
\begin{lstlisting}[linewidth=\linewidth,breaklines=true,breakatwhitespace=true,keepspaces=true]
[
  {
    "type": "function",
    "function": {
      "name": "segment_observer",
      "description": "Probe one interval with an MLLM under specified sampling.",
      "parameters": {
        "type": "object",
        "properties": {
          "interval": {
            "type": "object",
            "properties": {
              "start_sec": { "type": "number", "minimum": 0 },
              "end_sec":   { "type": "number", "exclusiveMinimum": 0 }
            },
            "required": ["start_sec", "end_sec"]
          },
          "query": { "type": "string" },
          "fps": { "type": "number", "exclusiveMinimum": 0, "default": 1 },
          "max_total_frames": { "type": "integer", "minimum": 1, "default": 32 }
        },
        "required": ["interval", "query"]
      }
    }
  },
  {
    "type": "function",
    "function": {
      "name": "stitched_observer",
      "description": "Probe multiple segments, stitch frames, then answer one question.",
      "parameters": {
        "type": "object",
        "properties": {
          "segments": {
            "type": "array",
            "minItems": 1,
            "items": {
              "type": "object",
              "properties": {
                "start_sec": { "type": "number", "minimum": 0 },
                "end_sec":   { "type": "number", "exclusiveMinimum": 0 },
                "fps":       { "type": "number", "exclusiveMinimum": 0, "default": 1 }
              },
              "required": ["start_sec", "end_sec"]
            }
          },
          "query": { "type": "string" },
          "global_interval": {
            "type": "object",
            "properties": {
\end{lstlisting}
    \end{tcolorbox}
    \caption{JSON Tool Schema used for \textit{Reasoner} of LensWalk (part 1). }
    \label{fig:tool-signatures}
\end{figure*}

\begin{figure*}[t]
    \ContinuedFloat
    \centering
    \begin{tcolorbox}[enhanced,breakable,colback=gray!10,colframe=gray!75,
                      arc=3mm,boxrule=0.5pt]
\linespread{0.8}\selectfont
\begin{lstlisting}[linewidth=\linewidth,breaklines=true,breakatwhitespace=true,keepspaces=true]
              "start_sec": { "type": "number", "minimum": 0 },
              "end_sec":   { "type": "number", "exclusiveMinimum": 0 }
            },
            "required": ["start_sec", "end_sec"]
          },
          "fps": { "type": "number", "exclusiveMinimum": 0, "default": 0.5 },
          "max_total_frames": { "type": "integer", "minimum": 1, "default": 128 }
        },
        "required": ["segments", "query"]
      }
    }
  },
  {
    "type": "function",
    "function": {
      "name": "scan_observer",
      "description": "Scan a global interval by slices and summarize each slice.",
      "parameters": {
        "type": "object",
        "properties": {
          "global_interval": {
            "type": "object",
            "properties": {
              "start_sec": { "type": "number", "minimum": 0 },
              "end_sec":   { "type": "number", "exclusiveMinimum": 0 }
            },
            "required": ["start_sec", "end_sec"]
          },
          "num_slices": { "type": "integer", "minimum": 1 },
          "slice_duration_sec": { "type": "number", "exclusiveMinimum": 0 },
          "query": { "type": "string" },
          "fps": { "type": "number", "exclusiveMinimum": 0, "default": 0.25 },
          "max_total_frames": { "type": "integer", "minimum": 1, "default": 180 }
        },
        "required": ["global_interval", "query"]
      }
    }
  },
  {
    "type": "function",
    "function": {
      "name": "finish",
      "description": "Return the final answer and end the dialog.",
      "parameters": {
        "type": "object",
        "properties": {
          "answer": { "type": "string" }
        },
        "required": ["answer"]
      }
    }
  }
]
\end{lstlisting}
    \end{tcolorbox}
    \caption{JSON Tool Schema used for \textit{Reasoner} of LensWalk (continued).}
\end{figure*}
\section{More Results}
\label{app:more results}
    In this section, we provide additional qualitative results and trajectory visualizations to complement the main experiments. All case studies are drawn from the long split of the Video-MME benchmark, and all trajectories are produced by the same o3 model acting as both the Reasoner and the Observer. All visualizations follow a unified annotation scheme. Within each reasoning trajectory, key pieces of evidence from visual observations that are crucial for driving the reasoner’s inference are highlighted in \textbf{\textcolor{red}{red}}. On the temporal axis, sampled contexts covered by different tools are indicated with colored intervals: \textbf{\textcolor[RGB]{174,218,145}{green}} for 
\textbf{\textcolor[RGB]{174,218,145}{ScanSearch}}, \textbf{\textcolor[RGB]{150,232,202}{blue}} for 
\textbf{\textcolor[RGB]{150,232,202}{Segment Focus}}
, and \textbf{\textcolor[RGB]{255,230,158}{yellow}} for 
\textbf{\textcolor[RGB]{255,230,158}{Stitched Verify}} respectively. 

% Insert after B.3 Baseline Reproduction and before \section{More Results}

\subsection{Additional Quantitative Analyses}
\label{sec:additional_quantitative}

\paragraph{Shared-backbone comparison and end-to-end cost.}
Table~\ref{tab:shared_o3_lvbench} compares LensWalk with prior video agents on LVBench under a shared o3 reasoner. LensWalk requires no offline captioning or index construction, so its end-to-end cost is substantially lower than retrieval-based agents once preprocessing is counted. It also uses 14--32$\times$ fewer frames per query while outperforming MR. Video and VideoAgent. Compared with DVD, LensWalk trades some accuracy for zero preprocessing and much lower total cost.
\begin{table*}[!t]
    \caption{Shared-backbone comparison on LVBench using o3 as the reasoner. We report accuracy, online inference time, offline preprocessing time, and average frames consumed per query. Retrieval-based agents require dense pre-processing of the full video, while LensWalk does not.}
    \label{tab:shared_o3_lvbench}
    \vspace{-4pt}

    \centering
    \setlength{\tabcolsep}{5pt}
    \renewcommand{\arraystretch}{0.92}
    \resizebox{0.97\linewidth}{!}{%
    \begin{tabular}{lcccc}
        \toprule
        \textbf{Method} & \textbf{Acc. (\%)} & \textbf{Online Inference (s)} & \textbf{Offline Preprocessing (s)} & \textbf{Avg. Frames / Query} \\
        \midrule
        o3         & 57.1 & 38.9   & 0      & 256 \\
        \midrule
        LensWalk   & 68.6 & 190.35 & 0      & 290.3 \\
        DVD        & 74.2 & 153.3  & 2180.4 & 8202 (2 fps) \\
        MR. Video  & 65.5 & 326.2  & 4135.2 & 9227 (2.25 fps) \\
        VideoAgent & 64.1 & 200.5  & 1131.3 & 4101 (1 fps) \\
        \bottomrule
    \end{tabular}}

    \vspace{2pt}
    \begin{minipage}{1\linewidth}
        \footnotesize\raggedright\color{gray}
        \textbf{Source.} Results for methods other than o3 and LensWalk are taken from DVD
        (\url{https://openreview.net/forum?id=oQYq9L1NVT}).
    \end{minipage}
    \vspace{-6pt}
\end{table*}
% \begin{table*}[!t]
%     \small
%     \centering
%     \caption[Shared-backbone comparison on LVBench using o3 as the reasoner.]%
% {Shared-backbone comparison on LVBench using o3 as the reasoner. We report accuracy, online inference time, offline preprocessing time, and average frames consumed per query. Retrieval-based agents require dense pre-processing of the full video, while LensWalk does not.}

%     \label{tab:shared_o3_lvbench}
%     \vspace{-6pt}
%     \setlength{\tabcolsep}{4pt}
%     \renewcommand{\arraystretch}{0.92}
%     \setlength{\aboverulesep}{0.2ex}
%     \setlength{\belowrulesep}{0.2ex}
%     \resizebox{\linewidth}{!}{%
%     \begin{tabular}{lcccc}
%         \toprule
%         \textbf{Method} & \textbf{Acc. (\%)} & \textbf{Online Inference (s)} & \textbf{Offline Preprocessing (s)} & \textbf{Avg. Frames / Query} \\
%         \midrule
%         o3         & 57.1 & 38.9   & 0      & 256 \\
%         \midrule
%         LensWalk   & 68.6 & 190.35 & 0      & 290.3 \\
%         DVD        & 74.2 & 153.3  & 2180.4 & 8202 (2 fps) \\
%         MR. Video  & 65.5 & 326.2  & 4135.2 & 9227 (2.25 fps) \\
%         VideoAgent & 64.1 & 200.5  & 1131.3 & 4101 (1 fps) \\
%         \bottomrule
%     \end{tabular}}
%     \vspace{-4pt}
% \end{table*}

\paragraph{Efficiency and adaptive budget allocation.}
Table~\ref{tab:adaptive_budget_scaling} shows that LensWalk does not impose a fixed reasoning overhead. It typically converges in 2.6--2.8 steps on short or simple benchmarks, but increases both interaction steps and frame usage as videos become longer or questions demand more reasoning. Together with Table~\ref{tab:shared_o3_lvbench}, this indicates that LensWalk improves efficiency not only by avoiding expensive full-video preprocessing, but also by allocating observation budget only when additional evidence is needed.

\begin{table*}[!t]
    \small
    \centering
    \caption{Adaptive scaling of LensWalk (o3/GPT-4.1) across benchmarks with different video lengths and reasoning demands. Gains are relative to the single-pass GPT-4.1 baseline.}
    \label{tab:adaptive_budget_scaling}
    \vspace{-6pt}
    \setlength{\tabcolsep}{3pt}
    \renewcommand{\arraystretch}{0.92}
    \setlength{\aboverulesep}{0.2ex}
    \setlength{\belowrulesep}{0.2ex}
    \resizebox{\linewidth}{!}{%
    \begin{tabular}{llcccccc}
        \toprule
        \textbf{Benchmark} & \textbf{Split} & \textbf{Acc. (\%)} & \textbf{Gain} & \textbf{Avg. Video Length (s)} & \textbf{Reasoning-Intensive} & \textbf{Avg. Steps} & \textbf{Avg. Frames Used} \\
        \midrule
        \multirow{3}{*}{Video-MME}
            & Short  & 82.3 & +1.7 & 80.7   & No  & 2.8 & 89.7 \\
            & Medium & 79.9 & +8.3 & 515.9  & No  & 4.2 & 233.0 \\
            & Long   & 70.0 & +6.9 & 2466.7 & No  & 6.8 & 387.1 \\
        EgoSchema
            & \textemdash & 74.8 & +2.6 & 180.0  & No  & 2.6 & 89.7 \\
        \midrule
        Video-MMMU
            & \textemdash & 77.1 & +9.7 & 506.2  & Yes & 4.8 & 178.4 \\
        MMVU
            & \textemdash & 80.9 & +2.0 & 51.4   & Yes & 3.1 & 58.5 \\
        \bottomrule
    \end{tabular}}
    \vspace{-4pt}
\end{table*}

\paragraph{Additional reasoners and a static extracted-frames baseline.}
Table~\ref{tab:additional_reasoners_static} separates the benefit of active observation from simply reusing better frames. Feeding the union of frames visited by LensWalk to a single-pass VLM yields only modest gains (+0.8 to +2.6 points), indicating that the improvement does not come from frame selection alone. The same table also shows that LensWalk remains effective with additional reasoners such as GLM-4.7 and DeepSeek-V3, indicating that the framework is not tied to a single planner.

\begin{table*}[!t]
    \small
    \centering
    \caption{Additional model recipes and a static extracted-frames baseline on Video-MME (long split). Each group header shows the corresponding single-pass observer baseline. \emph{Extracted Frames} feeds the union of frames visited by LensWalk under the same 256-frame budget into a single-pass VLM.}
    \label{tab:additional_reasoners_static}
    \vspace{-6pt}
    \setlength{\tabcolsep}{3pt}
    \renewcommand{\arraystretch}{0.92}
    \setlength{\aboverulesep}{0.2ex}
    \setlength{\belowrulesep}{0.2ex}
    \resizebox{\linewidth}{!}{%
    \begin{tabular}{lc lc lc}
        \toprule
        \multicolumn{2}{c}{\textbf{GPT-4.1 Observer (baseline: 63.1)}} &
        \multicolumn{2}{c}{\textbf{GPT-4.1-mini Observer (baseline: 59.4)}} &
        \multicolumn{2}{c}{\textbf{o3 Observer (baseline: 64.7)}} \\
        \cmidrule(lr){1-2}\cmidrule(lr){3-4}\cmidrule(lr){5-6}
        \textbf{Reasoner} & \textbf{Acc. (\%)} &
        \textbf{Reasoner} & \textbf{Acc. (\%)} &
        \textbf{Reasoner} & \textbf{Acc. (\%)} \\
        \midrule
        GLM-4.7                 & 68.3 (+5.2) & GLM-4.7                 & 66.0 (+6.6) & o3                         & 71.4 (+6.7) \\
        DeepSeek-V3             & 66.4 (+3.3) & DeepSeek-V3             & 63.3 (+3.9) & \textemdash                & \textemdash \\
        \emph{Extracted Frames} & 65.0 (+1.9) & \emph{Extracted Frames} & 62.0 (+2.6) & \emph{Extracted Frames}    & 65.5 (+0.8) \\
        \bottomrule
    \end{tabular}}
    % \vspace{-8pt}
\end{table*}
\subsection{Behaviour Analysis}
\begin{figure*}[t]
  \centering
  \includegraphics[width=0.8\linewidth, trim=0 10cm 0 0, clip]{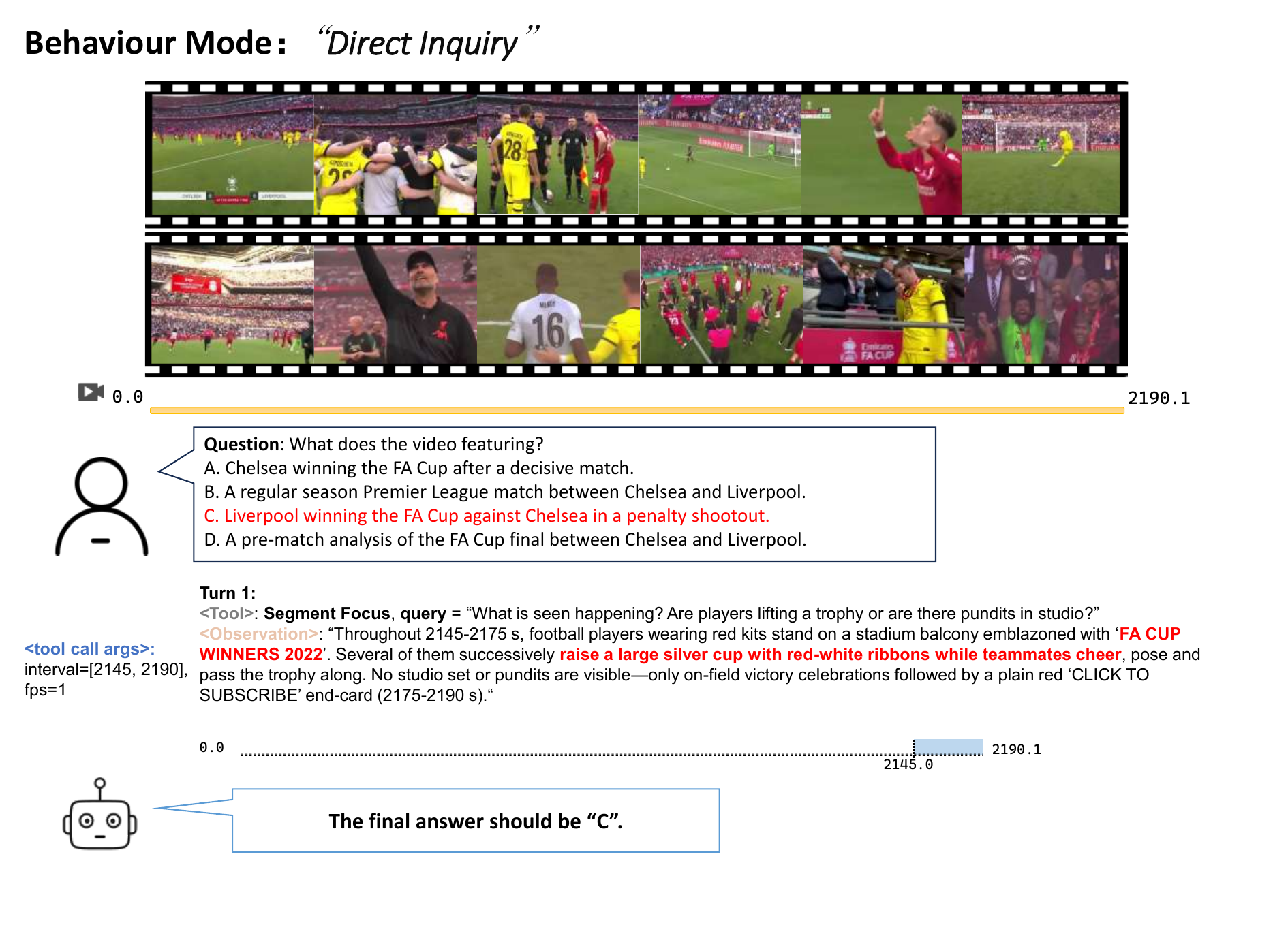}
  \caption{Example of \textbf{Direct Inquiry} behavior. }
  \label{fig:di_example}
\end{figure*}

Section \ref{analysis on tool-call} has introduced the taxonomy of tool-call behavior patterns; this section provides a detailed analysis of representative instances for each behavior.
\paragraph{Direct Inquiry.}
Direct Inquiry is the simplest tool-use pattern, where the agent can answer with only one or two tool calls. It is mainly used for straightforward temporal-localization questions such as “What happens at the beginning of the video?” or “What occurs between 10:21 and 10:25?”, as well as questions about the overall video type. As shown in Fig.~\ref{fig:di_example}, the agent only queries the ending interval for a specific option and then finds that it is correct.

\begin{figure*}[t]
  \centering
  \includegraphics[width=0.8\linewidth]{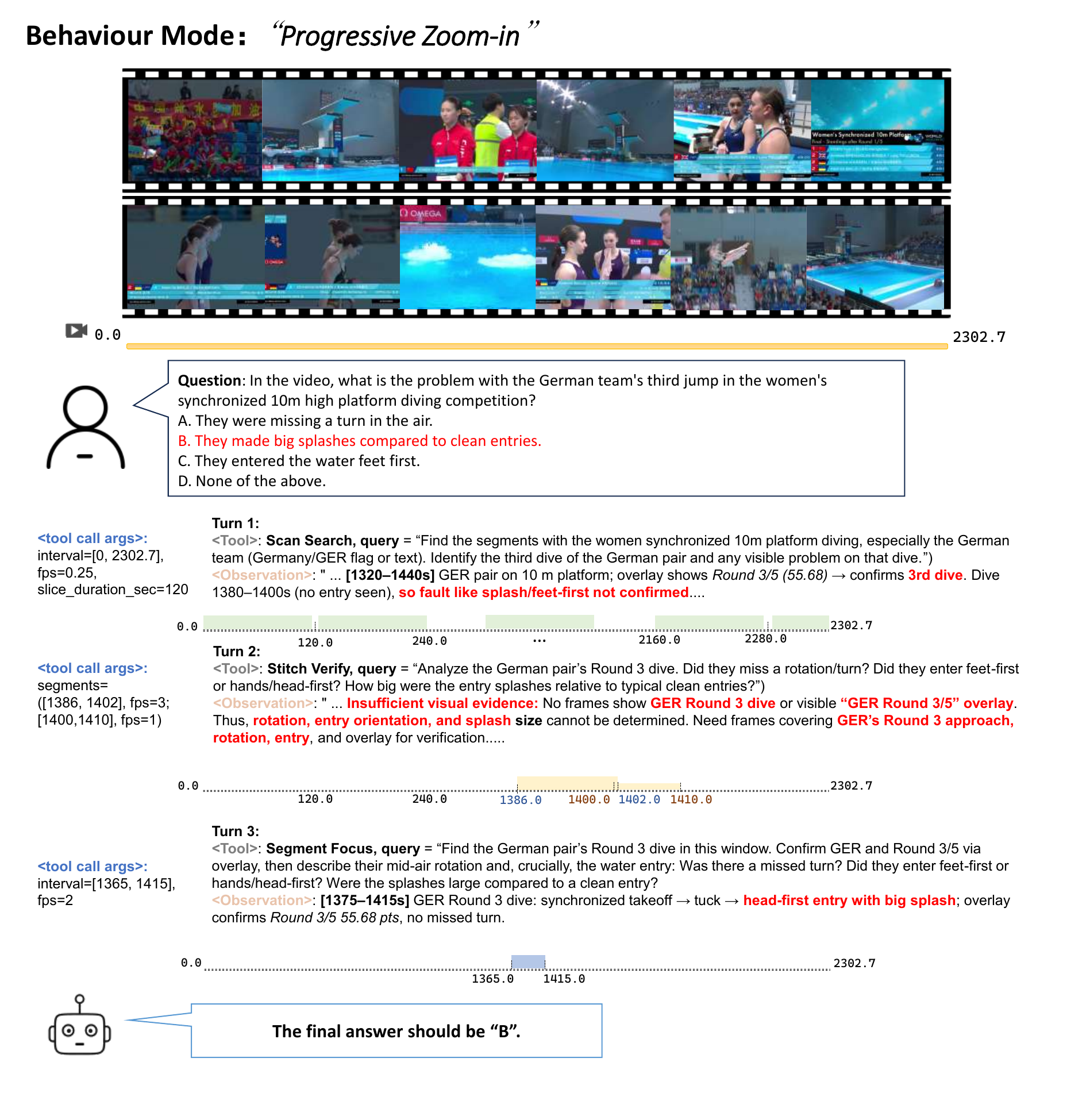}
  \caption{Example of \textbf{Progressive Zoom-in} behavior. }
  \label{fig:pz_example}
\end{figure*}

\paragraph{Progressive Zoom-in.}
The Progressive Zoom-in pattern begins with a coarse global scan and gradually narrows the candidate segments. Using approximate information from earlier steps, the agent then conducts increasingly fine-grained observations within a smaller temporal window. As illustrated in Fig.~\ref{fig:pz_example}, the model first performs a global \textit{ScanSearch} to locate intervals containing key visual cues mentioned in the question (e.g., the German team and the diving scene), and then progressively restricts its focus around the queried ``10m high platform diving competition'', eventually arriving at the correct answer.

\begin{figure*}[t]
  \centering
  \includegraphics[width=0.8\linewidth, trim=0 5cm 0 0, clip]{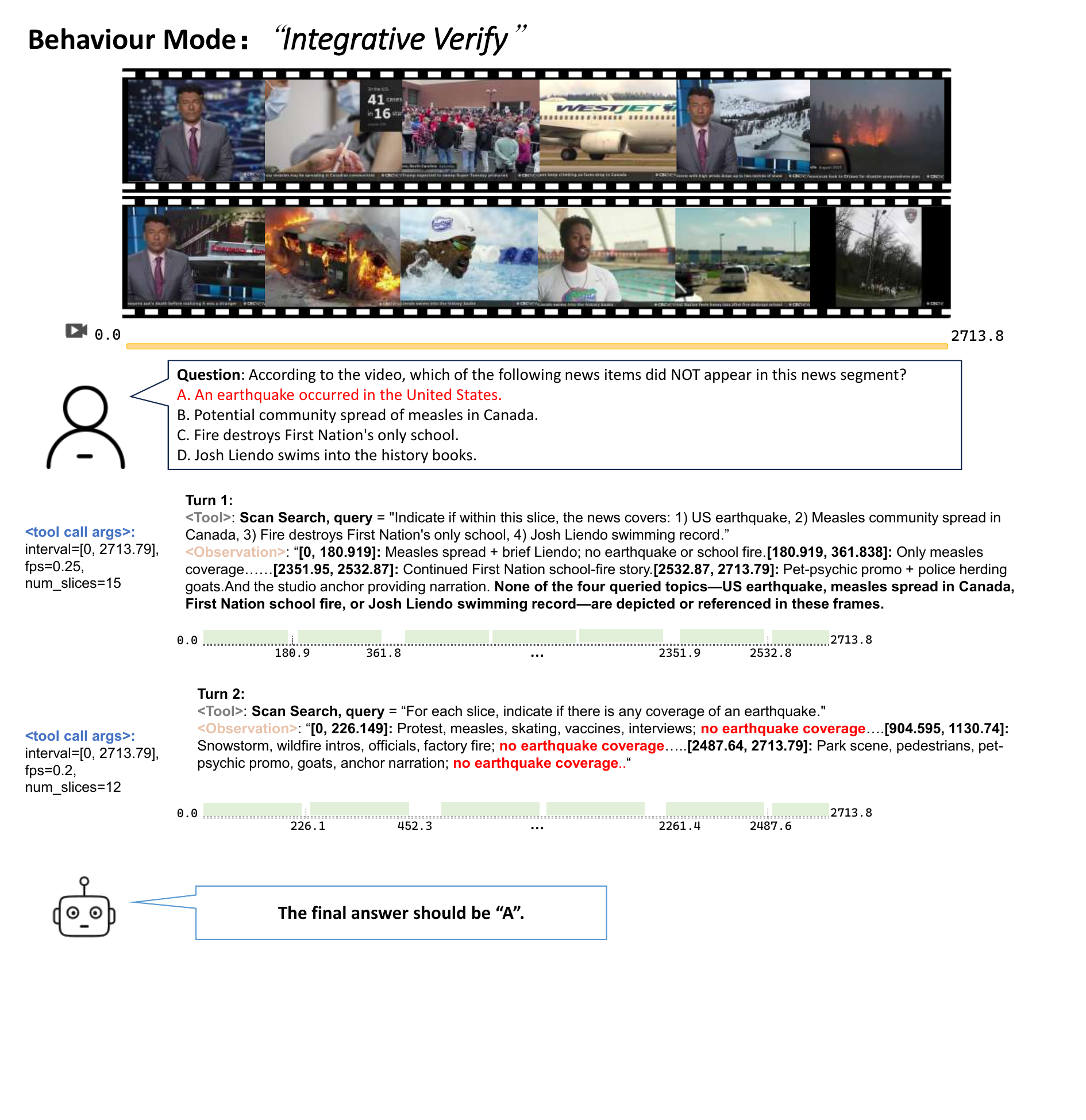}
  \caption{Example of \textbf{Integrative Verify} behavior. }
  \label{fig:iv_example}
\end{figure*}

\paragraph{Integrative Verify.}
In the Integrative Verify pattern, after several preliminary observations, the agent formulates an explicit hypothesis and then tests it by jointly inspecting multiple related segments. As illustrated in Fig.~\ref{fig:iv_example}, given a query such as ``Is there any segment where they say an earthquake occurred?'', the model systematically reviews each clip for mentions or visual evidence of an earthquake and, upon finding none, concludes that no such event is reported.

\begin{figure*}[t]
  \centering
  \includegraphics[width=0.8\linewidth, trim=0 7cm 0 0, clip]{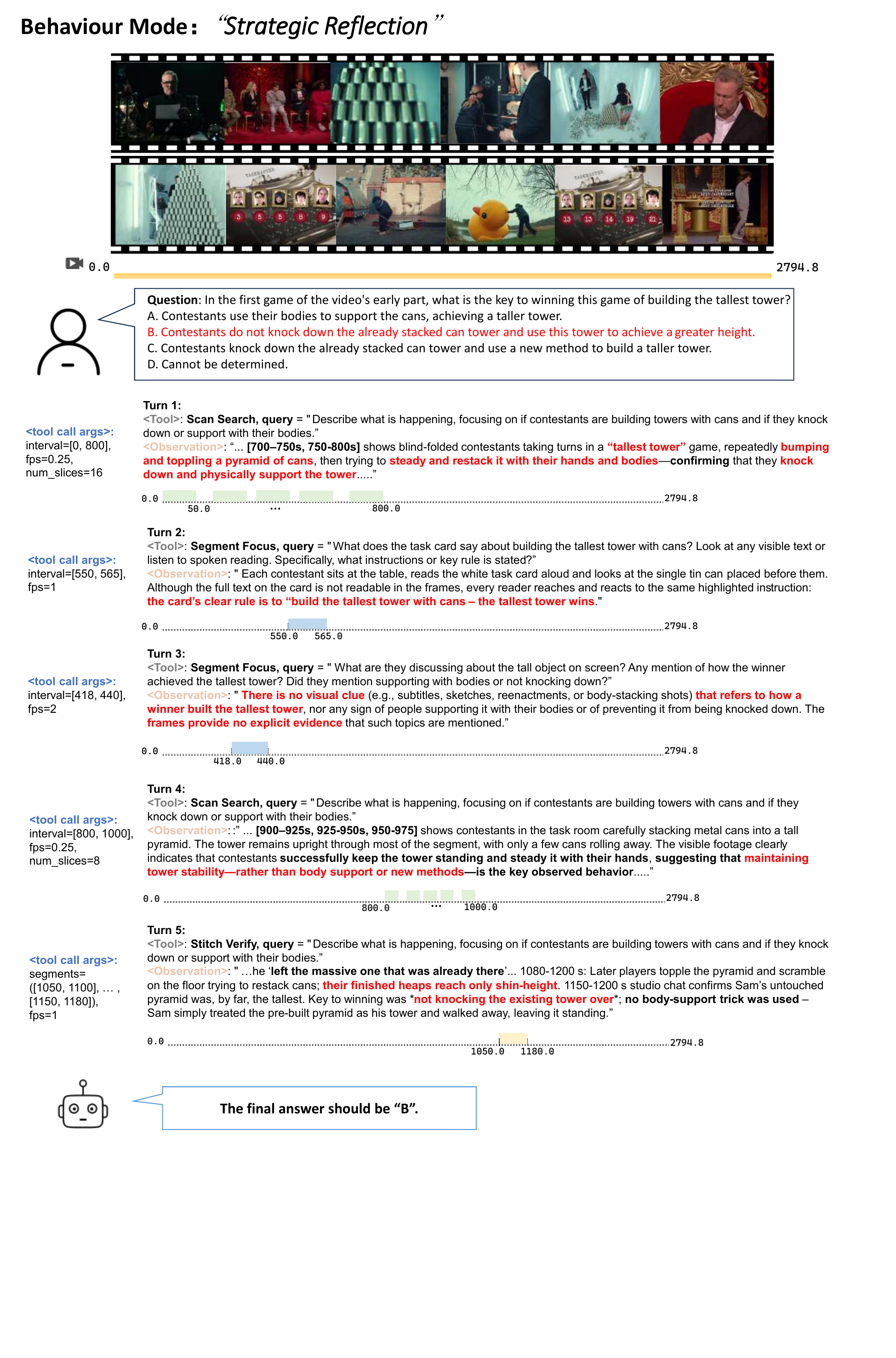}
  \caption{Example of \textbf{Strategic Reflection} behavior. }
  \label{fig:sr_example}
\end{figure*}

\paragraph{Strategic Reflection.}
Strategic Reflection is a higher-level reasoning pattern in which the agent realizes that its current observational strategy is failing, then backtracks to explore broader temporal regions before refining again, rather than over-focusing on an uninformative segment. As illustrated in Fig.~\ref{fig:sr_example}, once the model recognizes that the subtitles in the initial interval do not explain the game rules, it strategically shifts to another part of the video, enabling it to observe the optimal game strategy and ultimately obtain the correct answer.

\begin{figure*}[t]
  \centering
  \includegraphics[width=0.8\linewidth, trim=0 6cm 0 0, clip]{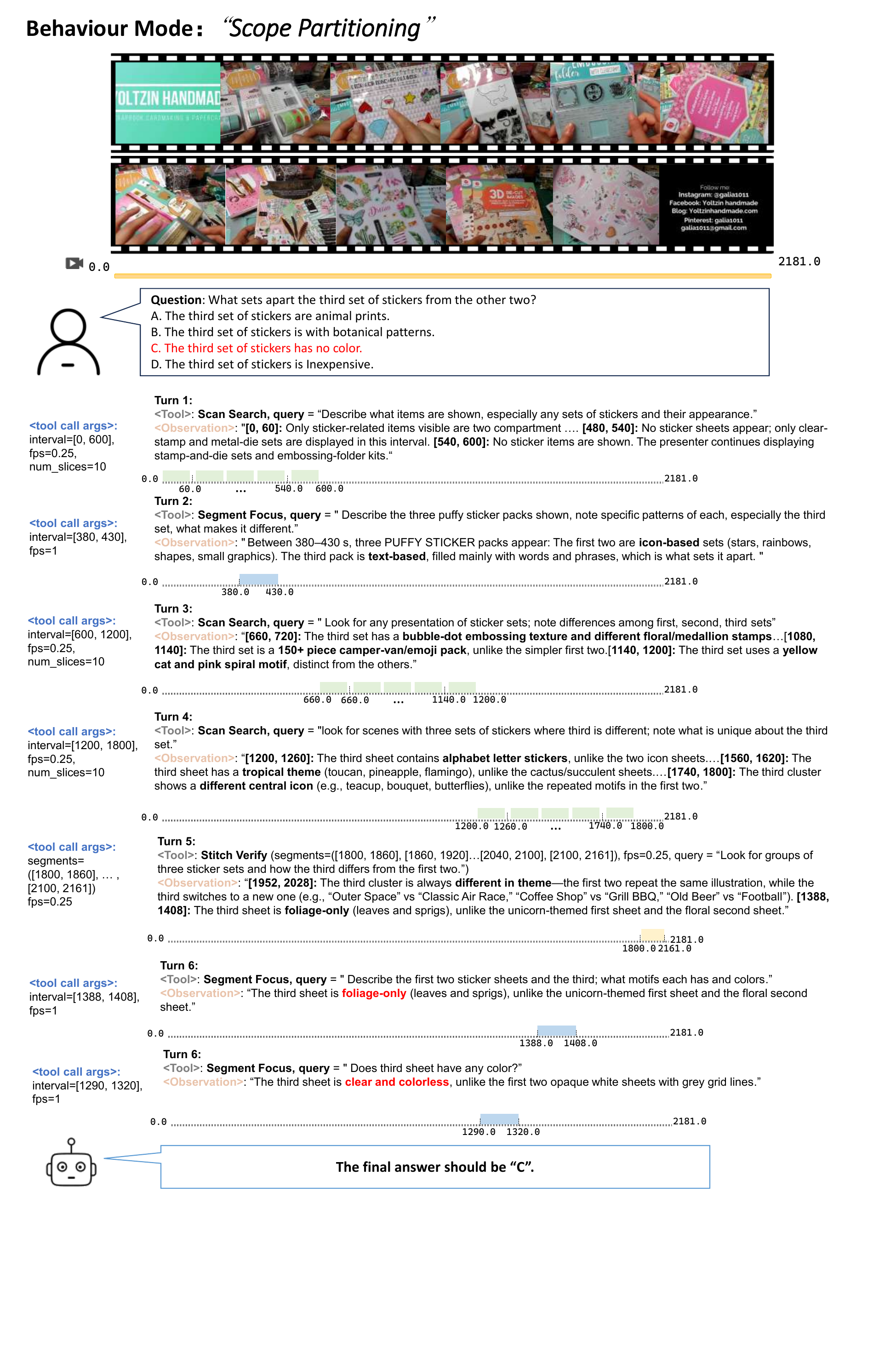}
  \caption{Example of \textbf{Scope Partitioning} behavior. }
  \label{fig:sp_example}
\end{figure*}

\paragraph{Scope Partitioning.}
Scope Partitioning adopts a divide-and-conquer strategy: the agent partitions the video into coarse segments and sequentially checks whether each region contains clues worth closer inspection. As shown in Fig. ~\ref{fig:sp_example}, it first scans the intervals 0--600s, 600--1200s, and 1200--1800s to locate the position of the clue about the third set, then identifies 1200--1800s as the most promising range, performs more fine-grained observations there, and finally derives the correct answer.

\begin{figure*}[t]
  \centering
  \includegraphics[width=0.8\linewidth, , trim=0 9cm 0 0, clip]{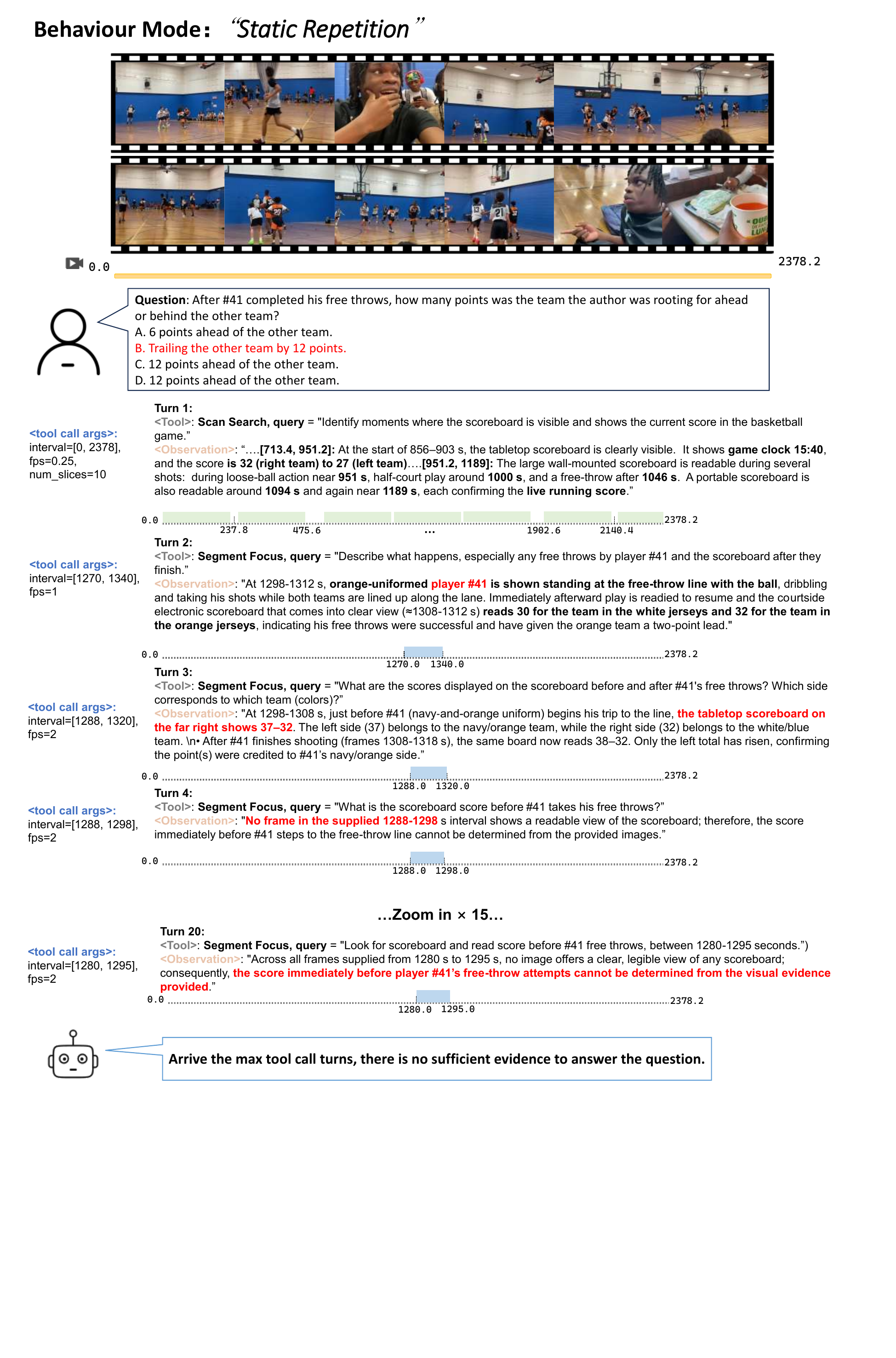}
  \caption{Example of \textbf{Static Repetition} behavior.  The notation “zoom in × n” indicates that several fine-grained but low-impact intermediate inspections are omitted from the visualization for brevity.}
  \label{fig:rp_example}
\end{figure*}

\paragraph{Static Repetition.}
In contrast to \emph{Strategic Reflection}, \emph{Static Repetition} denotes a rigid mode in which the model becomes trapped in an unproductive strategy. Rather than exploring alternative temporal regions or adjusting its observation queries, it repeatedly issues nearly identical observe calls with the same intent on the same segment, even after multiple failed attempts to find supporting evidence, as shown in Fig.~\ref{fig:rp_example}. This fixation on a single interval prevents the model from discovering the correct answer elsewhere in the video.

\subsection{Failure Modes and Analysis}
\begin{figure*}[t]
  \centering
  \includegraphics[width=0.8\linewidth, , trim=0 5cm 0 0, clip]{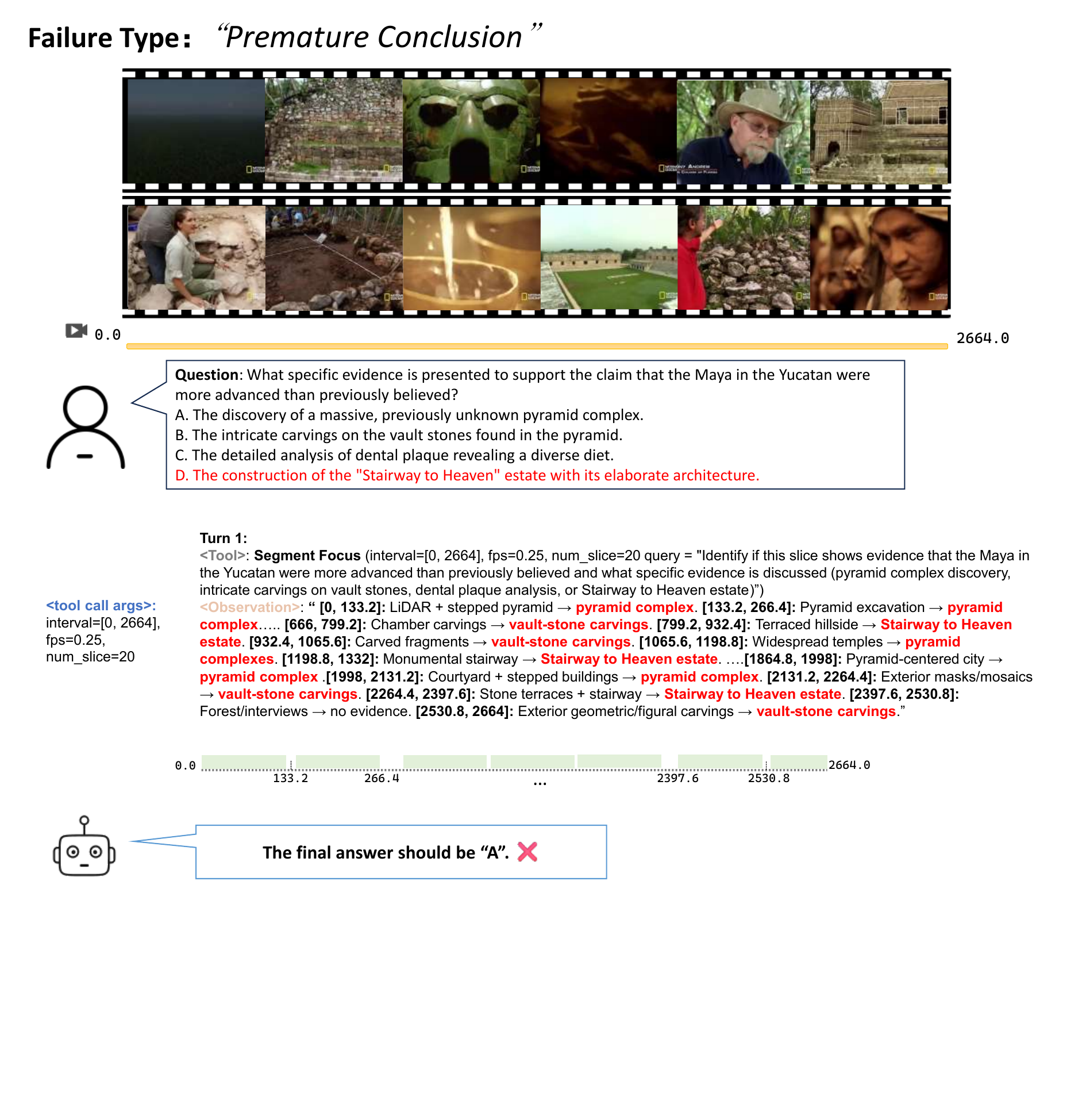}
  \caption{Example of Failure mode: \textbf{Premature Conclusion}. }
  \label{fig:fail_pc_example}
\end{figure*}

In this section, we illustrate a few failure modes observed during our evaluation. Analyzing these error patterns provides valuable insights into the limitations of current Large Language Models (LLMs) acting as reasoning planners and highlights critical directions for future active video agent design. We identify four primary failure patterns:

\noindent \textbf{Premature Conclusion.} The agent terminates the reasoning loop too early (often within 1–2 turns) upon encountering the most frequent cue “pyramid complex” in the initial observation, as shown in Fig.~\ref{fig:fail_pc_example}. By failing to validate this cue against potential distractors or cross-reference it with other video segments, the model exhibits unwarranted overconfidence. This failure mode suggests that future agents require a stronger internal “critic” or verification mechanism to balance token efficiency with rigorous hypothesis testing.

\begin{figure*}[t]
  \centering
  \includegraphics[width=0.8\linewidth, , trim=0 3cm 0 0, clip]{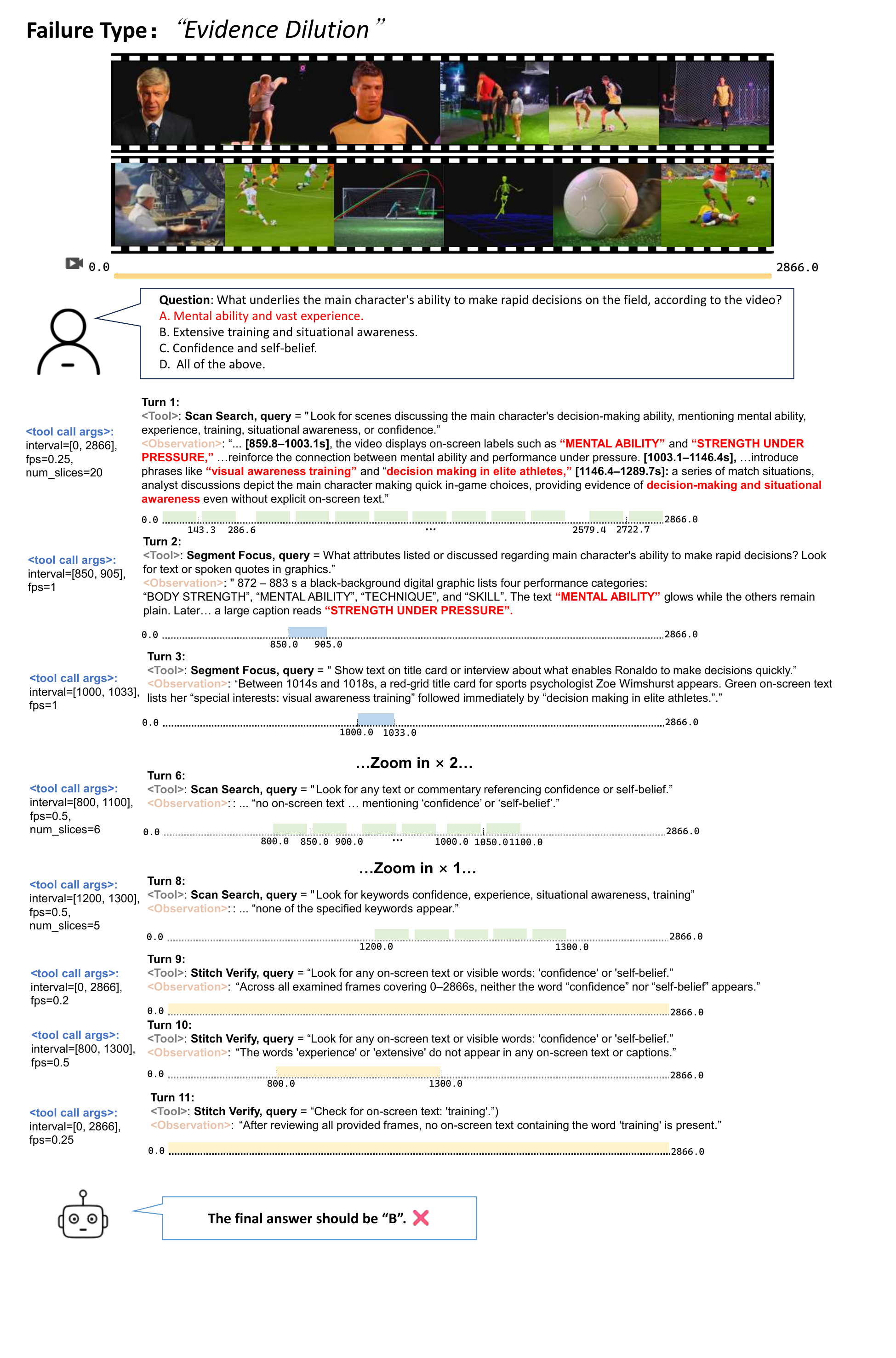}
  \caption{Example of Failure mode: \textbf{Evidence Dilution}.  The notation ``zoom in × n'' indicates that several fine-grained but low-impact intermediate inspections are omitted from the visualization for brevity.}
  \label{fig:fail_ed_example}
\end{figure*}

\noindent \textbf{Evidence Dilution.} Conversely, excessive observation can degrade performance. We observe cases, as in Fig.~\ref{fig:fail_ed_example}, where the agent discovers strong, decisive evidence in early turns e.g. ``STRENGTH UNDER PRESSURE'' or ``MENTAL ABILITY'', to the correct choice ``Mental ability and vast experience'', but continues to explore irrelevant segments. As the context window fills with ``loose'' or noisy information, the strong initial evidence gets diluted. The agent finally succumbs to recency bias or confusion, hallucinating an answer based on the latest weak observations. This underscores the importance of active memory management—agents must learn not just when to seek information, but when to prune irrelevant context to maintain reasoning clarity.

\begin{figure*}[t]
  \centering
  \includegraphics[width=0.8\linewidth, , trim=0 4cm 0 0, clip]{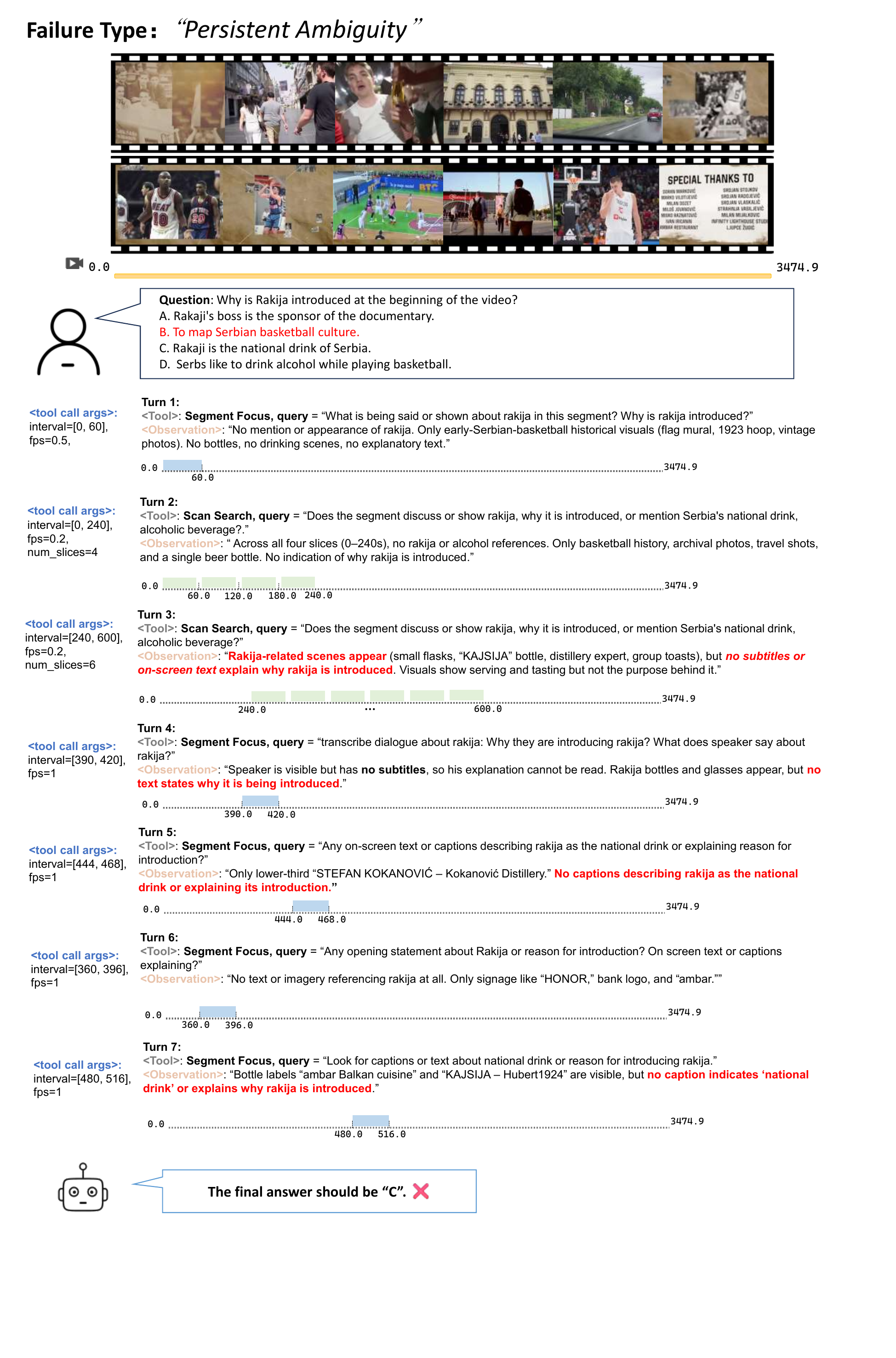}
  \caption{Example of Failure mode: \textbf{Persistent Ambiguity}. }
  \label{fig:fail_pa_example}
\end{figure*}

\noindent \textbf{Persistent Ambiguity.} The agent may persistently search along an uninformative trajectory and never obtain decisive visual evidence, ultimately being forced to give an ambiguous, semantics-only guess rather than a visually grounded answer. As shown in Fig. ~\ref{fig:fail_pa_example}, the agent keeps focusing on related subtitles instead of frame-wise visual changes, misses the key opening event, and therefore predicts incorrectly.

\noindent \textbf{Static Repetition.} Finally, as noted in the behavioral analysis, the agent may fall into a degenerate loop, repeatedly issuing identical tool calls for the same timestamp without adjusting its parameters. This reflects a failure in the agent's meta-cognition or state tracking; the planner fails to recognize that its current strategy is unproductive, highlighting the need for mechanisms that detect stagnation and trigger strategic shifts (e.g., from \emph{Segment Focus} to \emph{Strategic Reflection}).
% \subsubsection{Failure Modes}
% Beyond the previously discussed \emph{Static Repetition} pattern, which frequently leads to failure, we also identify two additional failure modes. 

% First, after observing a single salient visual cue, the model may prematurely assume that this cue constitutes all the essential evidence required to answer the question, resulting in an unwarranted or overconfident conclusion. 

% Second, for more complex queries, the model may, after multiple rounds of exploration, determine that the available visual information is still insufficient for a definitive answer (e.g., missing textual cues or incomplete context in the video). In such cases, the model is unable to conclusively verify all answer options and is forced to rely on partial evidence, producing an inherently ambiguous or uncertain response.
% appendix A: Method Details
% A.1 Prompt
% A.2 Tool Format and Signature
% B: Evaluation Details
% B.1 Model setting
% B.2 Dataset Setting
% C: More Results
%C.2: Behaviour Analysis
% WARNING: do not forget to delete the supplementary pages from your submission 
% \input{sec/X_suppl}
x

\end{document}